\documentclass[10pt,twocolumn,letterpaper]{article}

\usepackage[pagenumbers]{cvpr} 

\usepackage{graphicx}
\usepackage{amsmath}
\usepackage{amssymb}
\usepackage{booktabs}
\usepackage{esvect}
\usepackage{tabularx}
\usepackage{epigraph}
\usepackage{xcolor,colortbl}
\usepackage{lscape}
\usepackage{rotating}
\usepackage{epstopdf}

\definecolor{lavender}{rgb}{0.9, 0.9, 0.98}
\newcolumntype{L}[1]{>{\raggedright\let\newline\\\arraybackslash\hspace{0pt}}m{#1}}
\newcolumntype{C}[1]{>{\centering\let\newline\\\arraybackslash\hspace{0pt}}m{#1}}
\newcolumntype{R}[1]{>{\raggedleft\let\newline\\\arraybackslash\hspace{0pt}}m{#1}}

\newcommand{\smallsec}[1]{\vspace{0.2em}\noindent\textbf{#1}}



\usepackage[pagebackref,breaklinks,colorlinks]{hyperref}

\usepackage[capitalize]{cleveref}
\crefname{section}{Sec.}{Secs.}
\Crefname{section}{Section}{Sections}
\Crefname{table}{Table}{Tables}
\crefname{table}{Tab.}{Tabs.}

\begin{document}

\title{Breaking the “Object” in Video Object Segmentation}

\author{\text{\qquad Pavel Tokmakov} \text{\qquad Jie Li} \text{\qquad Adrien Gaidon}\\
Toyota Research Institute\\
{\tt\small first.last@tri.global}
}

\maketitle

\begin{abstract}
The appearance of an object can be fleeting when it transforms. As eggs are broken or paper is torn,  their color, shape and texture can change dramatically, preserving virtually nothing of the original except for the identity itself. Yet, this important phenomenon is largely absent from existing video object segmentation (VOS) benchmarks. In this work, we close the gap by collecting a new dataset for Video Object Segmentation under Transformations (VOST). It consists of more than 700 high-resolution videos, captured in diverse environments, which are 21 seconds long on average and densely labeled with instance masks. We adopt a careful, multi-step approach to ensure that these videos focus on complex object transformations, capturing their full temporal extent. We then extensively evaluate state-of-the-art VOS methods and make a number of important discoveries. In particular, we show that existing methods struggle when applied to this novel task and that their main limitation lies in over-reliance on static appearance cues. This motivates us to propose a few modifications for the top-performing baseline that improve its capabilities by better modeling spatio-temporal information. More broadly, our work highlights the need for further research on learning more robust video object representations.
\end{abstract}

 \begin{flushright}
 \fontsize{10}{12}
 \textit{Rien ne se perd, rien ne se crée, tout se transforme.~~~} \\[1ex]
 \fontsize{10}{12}\selectfont
 Antoine Lavoisier\\
 \end{flushright}

\section{Introduction}
\label{sec:intro}
Spatio-temporal cues are central in segmenting and tracking objects in humans, with static appearance playing only a supporting role~\cite{hollingworth2009object,kahneman1992reviewing,scholl2007object}. In the most extreme scenarios, we can even localize and track objects defined by coherent motion alone, with no unique appearance whatsoever~\cite{gao2010objects}. Among other benefits, this appearance-last approach increases robustness to sensory noise and enables object permanence reasoning~\cite{peters2021capturing}. 
By contrast, modern computer vision models for video object segmentation~\cite{seong2020kernelized,yang2021associating,cheng2022xmem,athar2022hodor} operate in an appearance-first paradigm. Indeed, the most successful approaches effectively store patches with associated instance labels and retrieve the closest patches to segment the target frame~\cite{oh2019video,seong2020kernelized,yang2021associating,cheng2022xmem}. 

\begin{figure}[t]
\begin{center}
  \includegraphics[width=1\columnwidth]{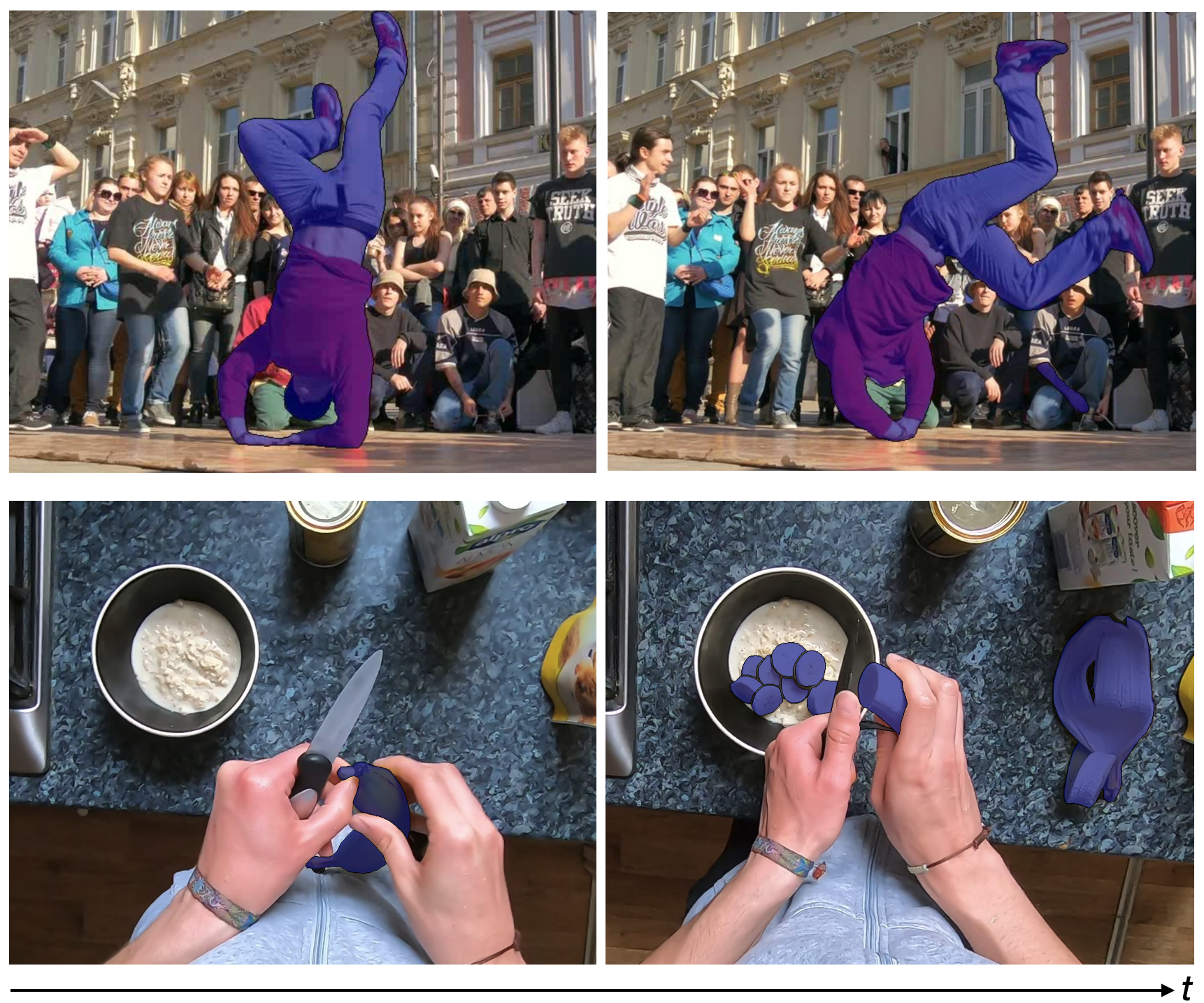}
\end{center}
\vspace{-5mm}
  \caption{
  Video frames from the DAVIS'17 dataset~\cite{pont20172017} (above), and our proposed VOST (below). While existing VOS datasets feature many challenges, such as deformations and pose change, the overall appearance of objects varies little. Our work focuses on object transformations, where appearance is no longer a reliable cue and more advanced spatio-temporal modeling is required. 
  }
\vspace{-3mm}
\label{fig:teaser}
\end{figure}

What are the reasons for this stark disparity? While some are algorithmic (e.g.,~object recognition models being first developed for static images), a key reason lies in the datasets we use. See for instance the “Breakdance” sequence from the validation set of DAVIS'17~\cite{pont20172017} in Figure~\ref{fig:teaser}: while the dancer's body experiences significant deformations and pose changes, the overall appearance of the person remains constant, making it an extremely strong cue. 
\begin{figure*}[t]
    \centering
    \includegraphics[width=1.\linewidth]{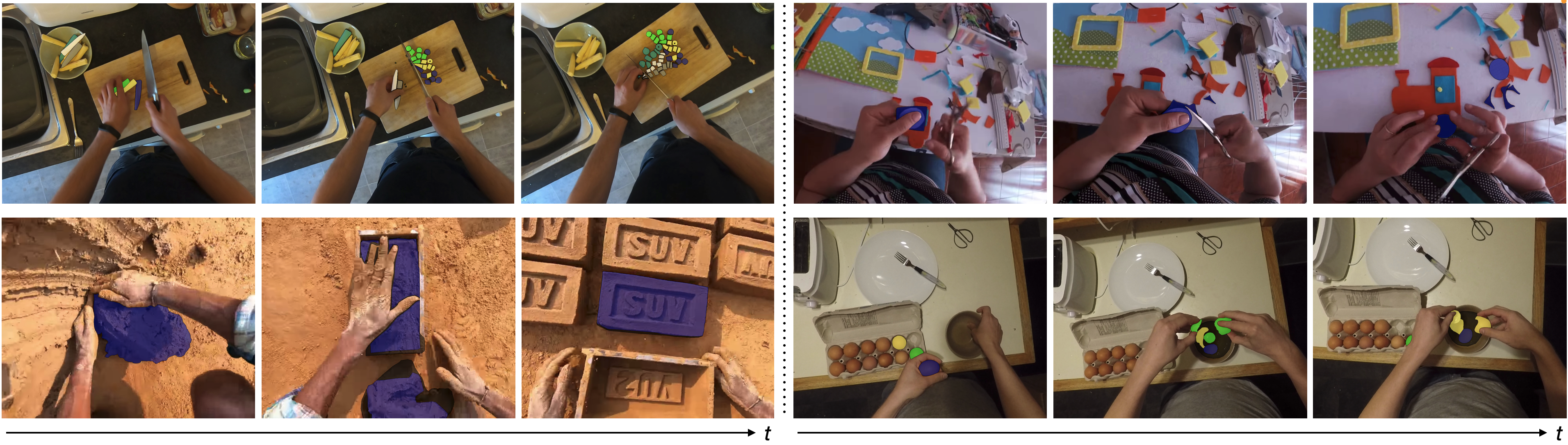}
    \vspace{-6mm}
    \caption{Representative samples from VOST with annotations at three different time steps (see \href{https://youtu.be/HvNRibYKip0}{video} for full results). Colours indicate instance ids, with grey representing ignored regions. VOST captures a wide variety of transformations in diverse environments and provides pixel-perfect labels even for the most challenging sequences. }
    \label{fig:samples}
    \vspace{-3mm}
\end{figure*}

However, this example -- representative of many VOS datasets -- covers only a narrow slice of the life of an object. In addition to translations, rotations, and minor deformations, objects can transform. Bananas can be peeled, paper can be cut, clay can be molded into bricks, etc. These transformations can dramatically change the color, texture, and shape of an object, preserving virtually nothing of the original except for the identity itself (see Figure~\ref{fig:teaser}, bottom and Figure~\ref{fig:samples}).
As we show in this paper, tracking object identity through these changes is relatively easy for humans (e.g. labelers), but very challenging for VOS models.
In this work, we set out to fill this gap and study the problem of segmenting objects as they undergo complex transformations. 

We begin by collecting a dataset that focuses on these scenarios in Section~\ref{sec:data}. We capitalize on the recent large-scale, ego-centric video collections~\cite{grauman2022ego4d,damen2022rescaling}, which contain thousands of examples of human-object interactions with activity labels. We carefully filter these clips to only include major object transformations using a combination of linguistic cues (change of state verbs~\cite{levin1993english,fillmore1967grammar}) and manual inspection. The resulting dataset, which we call VOST (Video Object Segmentation under Transformations), contains 713 clips, covering 51 transformations over 155 object categories with an average video length of 21.2 seconds. We then densely label these videos with more than 175,000 masks, using an unambiguous principle inspired by spatio-temporal continuity: if a region is marked as an object in the first frame of a video, all the parts that originate from it maintain the same identity (see Figure~\ref{fig:samples}).

Equipped with this unique dataset, we analyze state-of-the-art VOS algorithms in Section~\ref{sec:anal}. We strive to include a representative set of baselines that illustrates the majority of the types of approaches to the problem in the literature, including classical, first frame matching methods~\cite{Yang2018osmn}, local mask-propagation objectives~\cite{jabri2020space}, alternative, object-level architectures~\cite{athar2022hodor}, and the mainstream memory-based models~\cite{yang2020collaborative,yang2021associating,yang2021collaborative,cheng2022xmem}. Firstly, we observe that existing methods are indeed ill-equipped for segmenting objects through complex transformations, as illustrated by the large (2.3-12.5 times) gap in performance between VOST and DAVIS'17 (see Table~\ref{tab:sota}). A closer analysis of the results reveals the following discoveries: (1) performance of the methods is inversely proportional to their reliance on static appearance cues; (2) progress on VOST can be achieved by improving the spatio-temporal modeling capacity of existing architectures; (3) the problem is not easily solvable by training existing methods on more data.   

We conclude in Section~\ref{sec:disc} by summarizing the main challenges associated with modeling object transformations. We hope that this work will motivate further exploration into more robust video object representations. Our dataset, source code, and models are available at \href{https://vostdataset.org}{vostdataset.org}. 

\section{Related Work}
In this work, we study the problem of \textit{video object segmentation} under transformations and analyze existing \textit{VOS methods} under this novel task. Our efforts are motivated by observations about \textit{object perception in humans}. Below, we review the most relevant works on each of these topics.  
\begin{table*}[t]
  \centering
   \resizebox{\linewidth}{!}{
  \begin{tabular}{lccccccc}
    \toprule
    Dataset &
    \begin{tabular}{c}Videos\end{tabular} &
    \begin{tabular}{c}Frames\end{tabular} &
    \begin{tabular}{c}Avg len. (s)\end{tabular} &
    \begin{tabular}{c}Masks/frame\end{tabular} &
    \begin{tabular}{c}Ann fps.\end{tabular} &
    \begin{tabular}{c}Granularity\end{tabular} &
    \begin{tabular}{c}Focus\end{tabular}  \\\midrule
    DAVIS'16~\cite{perazzi2016benchmark}      & 50    & 3,455 & 3.0s & 1.0 & 24  & Binary & Data quality   \\
    DAVIS'17~\cite{pont20172017}    & 150 & 10,700 & 3.0s & 3.0 & 24 & Instance & Instance labels \\
    YTVOS~\cite{xu2018youtube}    & 3,500 & 120,400 & 4.6s & 1.6 & 6 & Instance & Dataset size \\
    UVO$^{*}$~\cite{wang2021unidentified}    & 10,337 & 30,500 & 3.0s & 8.8 & 1 & Instance & Object vocabulary \\
    VISOR~\cite{darkhalil2022epic}    & 7836 & 50,700 & 12.0s & 5.3 & 0.5 & Semantic & Object manipulation$^\dag$ \\
    VOST (Ours) & 713 & 75,547 & 21.2s & 2.3 & 5 & Instance & Object transformation\\
    \bottomrule
  \end{tabular}
   }
     \caption{Statistics of major video object segmentation datasets (*: train/val public annotations; \dag: including a small fraction of object transformation annotations). Unlike all existing VOS benchmarks, VOST focuses on the specific challenge of modeling complex object transformations. This motivates our design decisions to densely label relatively long videos with instance masks.
   }
  \label{tab:dataset_stats}
  \vspace{-3mm}
\end{table*}

\smallsec{Video object segmentation} is defined as the problem of pixel-accurate separation of foreground objects from the background in videos~\cite{tsai2012motion,li2013video,perazzi2016benchmark}. What constitutes foreground is either defined by independent motion~\cite{brox2010object,perazzi2016benchmark} or using a mask manually provided in the first frame of a video~\cite{tsai2012motion,li2013video,perazzi2016benchmark}, the latter setting known as semi-supervised VOS. The earliest datasets lacked in scale and consistency~\cite{brox2010object,tsai2012motion,li2013video}. The release of the DAVIS benchmark~\cite{perazzi2016benchmark} was a significant step for the community as it provided 50 high-resolution sequences featuring a variety of challenges. While DAVIS caused a flurry of novel VOS methods~\cite{caelles2017one,perazzi2017learning,tokmakov2017learning,voigtlaender2017online}, it treated VOS as a binary foreground/background separation problem.

In contrast DAVIS'17~\cite{pont20172017} not only extended the dataset to 150 videos, but, most importantly, introduced instance labels. In this, now de-facto standard, setting, an algorithm is provided with several object masks in the first frame and has to output pixel-perfect masks for these objects for the remainder of the video, together with their identity. While DAVIS focused on the data quality, it lacked in quantity, forcing most methods to resort to pre-training on static images~\cite{perazzi2017learning,khoreva2019lucid}, or synthetic videos~\cite{tokmakov2017learning}. This issue was addressed by the large-scale YouTube-VOS benchmark~\cite{xu2018youtube}, which features 3,252 videos over 78 categories.

Very recently, to further scale the datasets while keeping the annotation costs manageable, several works proposed to label videos at a very low fps (1 in~\cite{wang2021unidentified} and \(\sim\)0.5 in~\cite{darkhalil2022epic}) and interpolate ground truth labels to obtain dense annotations. The estimated labels are then automatically filtered to keep only the confident interpolations. While this approach was shown to work well in many cases, in Section~\ref{sec:interp} in the appendix we demonstrate that it fails precisely in the most challenging scenarios which we are interested in.

Notably, none of these datasets features a significant amount of object transformations. Thus, our effort is complementary to existing work. We compare VOST to major VOS benchmarks in Table~\ref{tab:dataset_stats}, illustrating our key design decisions. In particular, we label relatively long videos to capture the full extent of each transformation, and provide temporally dense instance-level labels, as interpolation fails when objects transform.

\smallsec{VOS methods} can be categorized in many possible ways. Here we focus on the semi-supervised setting and trace the history of the field to identify main trends. Early, pre-deep learning methods propagate the first frame labels over a spatio-temporal graph structure by optimizing an energy function~\cite{grundmann2010efficient,avinash2014seamseg,fan2015jumpcut}, but struggle with generalization due to their heuristic-based nature. 

First deep-learning solutions had to deal with the lack of video data for training and hence modeled video segmentation as an image-level problem~\cite{caelles2017one,xiao2018monet,khoreva2019lucid}. In particular, these works proposed to pre-train a CNN for binary object segmentation on COCO~\cite{lin2014microsoft} and then fine-tuned the model for a few iterations on the first frame of a test video. While this approach outperformed heuristic-based methods, it is computationally expensive and not robust to appearance change. These issues were separately addressed in~\cite{Yang2018osmn,chen2018blazingly,hu2018videomatch}, which replace expensive fine-tuning with cheap patch-level matching, and in~\cite{voigtlaender2017online,perazzi2017learning,luiten2018premvos,li2020delving} which introduce online adaptation mechanisms.

More recently, memory-based models have become the mainstream approach for semi-supervised video object segmentation~\cite{oh2018fast,oh2019video,voigtlaender2019feelvos,yang2020collaborative,seong2020kernelized,yang2021collaborative,yang2021associating,cheng2022xmem}. The earliest methods in this category~\cite{oh2018fast,voigtlaender2019feelvos,yang2020collaborative} extend the first-frame matching mechanism of~\cite{Yang2018osmn,chen2018blazingly,hu2018videomatch} by additionally matching with the previous frame. This architecture can be seen as a memory module with capacity 2, providing an efficient mechanism for adapting to appearance changes. More advanced versions of the architecture include increasing the memory capacity by storing several previous frames~\cite{oh2019video,seong2020kernelized}, using transformers~\cite{vaswani2017,devlin2018bert} for retrieving object labels from memory~\cite{yang2021associating,duke2021sstvos}, introducing memory compression to support longer sequences~\cite{cheng2022xmem,liang2020video}, and improving the efficiency of the memory read operation~\cite{xie2021efficient,cheng2021rethinking,seong2021hierarchical}.

Alternative approaches to VOS include supervised~\cite{perazzi2017learning,chen2020state,hu2017maskrnn} and, more recently, unsupervised~\cite{jabri2020space,wang2019learning} mask propagation methods that do not maintain an appearance model of the target. These methods are very efficient, but cannot handle occlusions and suffer from drift in longer sequences. A few works~\cite{liang2021video,zeng2019dmm,athar2022hodor} propose to perform appearance matching on the object, not on the patch level, but their accuracy remains low. Finally, coherent motion is a key signals for object perception in humans, but it was mostly studied in unsupervised VOS~\cite{tokmakov2017learning, yang2019unsupervised, yang2021self}.

\begin{figure*}[t]
    \centering
    \includegraphics[width=1\linewidth]{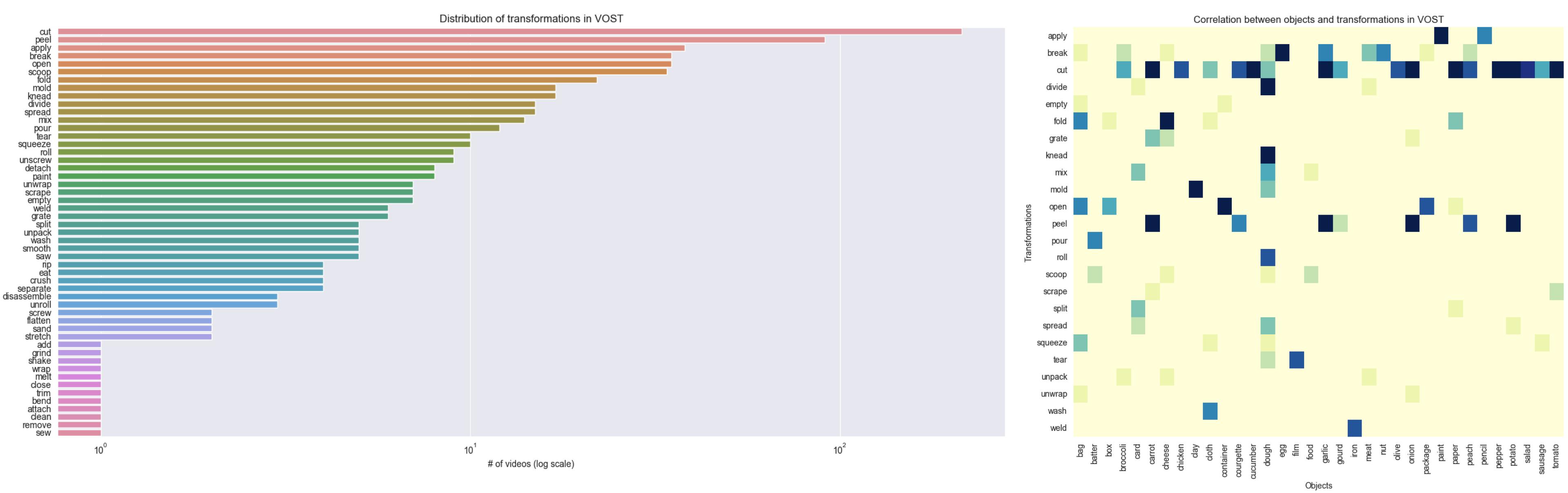}
    \vspace{-5mm}
    \caption{Statistics of VOST: distribution of transformations on the left, and co-occurrence statistics between the most common transformations and object categories on the right. While there is some bias towards common activities, like cutting, the tail of the distribution is sufficiently heavy. Moreover, cutting has a broad semantic meaning, resulting in diverse transformations. Best viewed with zoom.}
    \label{fig:stats}
    \vspace{-3mm}
\end{figure*}

In this work, we evaluate a representative set of semi-supervised VOS methods on the task of segmenting objects as they undergo complex transformations. Our experiments illustrate limitations of the appearance-first paradigm, motivating the exploration of spatio-temporal architectures. 

\smallsec{Object perception in humans} is driven by spatio-temporal cohesion. At the early development stages, infants use the notions of boundedness and cohesion in space-time, not static, gestalt cues like shape or texture to group surfaces into objects~\cite{spelke2013perceiving,spelke1990principles,spelke2007core}. In adults, the object files theory~\cite{kahneman1992reviewing} postulates that our visual system \textit{individuates} each object by grouping visual primitives based on spatio-temporal factors. Most importantly, object's individuation precedes its appearance identification, as shown in~\cite{hollingworth2009object,kahneman1992reviewing,scholl2007object}. That is, humans can perceive something as the same `thing' while its appearance remains in flux and might dramatically change over time. In the most extreme cases, individuation can function in the absence of any unique object appearance, as shown by Gao and Sholl~\cite{gao2010objects}.

Very recently, Peters and Kriegeskorte~\cite{peters2021capturing} summarized the differences between object representations in the brain and neural networks, including the dichotomy between spatio-temporal and appearance cues. They then argue that the best way to bridge the differences between these two types of representations is by introducing novel machine vision tasks that require more complex spatio-temporal reasoning. In this work, we make a step in this direction by extending the setting of video object segmentation to support object transformations. 

\section{Dataset Design and Collection}
\label{sec:data}
In this section, we discuss our approach to collecting VOST. The key steps include selecting representative videos, annotating them with instance masks, and defining an evaluation protocol. 

\subsection{Video selection}
\label{sec:select}
We choose to source our videos from the recent large-scale, egocentric action recognition datasets, which provide temporal annotations for a large vocabulary of activities. In particular, we use EPIC-KITCHENS~\cite{damen2022rescaling} and Ego4D~\cite{grauman2022ego4d}, where the former captures activities in kitchens, such as cooking or cleaning, and the later provides a much larger diversity of scenarios, including outdoor ones. It is worth noting that the egocentric focus of VOST is merely an artifact of the datasets that were used to source the videos. The nature of the problem itself is independent of the camera viewpoint and we expect that approaches developed on VOST will generalize to third-person videos.

While these datasets feature tens of thousands of clips, the vast majority of the actions (e.g., `take' or `look') do not result in object transformations. To automatically filter out such irrelevant clips, we capitalize on the notion of change of state verbs from the language theory~\cite{levin1993english,fillmore1967grammar}. That is, rather than manually filtering the videos themselves, we first filter the action labels. This dramatically reduces the total number of clips we have to consider to 10,706 (3,824 from EPIC-KITCHENS and 6,882 from Ego4D). 

Although all the clips selected above feature an object state change, not all result in a significant appearance change. For example, folding a towel in half, or shaking a paintbrush does nearly nothing to their overall appearance. To focus on the more challenging scenarios, we manually review each video and label its complexity on a scale from 1 to 5, where 1 corresponds to no visible object transformation and 5 to a major change of appearance, shape and texture (see Section~\ref{sec:compcat} in the appendix for details). In addition, at this stage we merge clips representing several steps of the same transformation (e.g. consecutive cuts of an onion). After collecting these labels we find that the majority of videos in the wild are not challenging, however, we are still left with 986 clips in the 4-5 range, capturing the entire temporal extent of these complex transformations.

Finally, we further filter the clip based on two criteria. Firstly, some videos are nearly impossible to label accurately with dense instance masks (e.g., due to excessive motion blur), so we skip them. Secondly, there are a few large clusters of near duplicates (e.g., there are 116 clips of molding clay into bricks that are performed by the same actor in the same environment), so we sub-sample those to reduce bias. The resulting dataset contains 713 videos covering 51 transformations over 155 object categories. Note that, in accordance with the standard VOS protocols~\cite{pont20172017,xu2018youtube}, semantic labels are only used for data collection and are not provided as input to the algorithms. 

The distribution over transformations and co-occurrence statistics between transformations and objects are shown in Figure~\ref{fig:stats}. Firstly we observe that, although there is some bias towards more common actions, such as cutting, the long tail of interactions is sufficiently heavy. Moreover, as evident from the correlation statistics on the right side of the figure, cutting has an extremely broad semantic meaning and can be applied to almost any object, resulting in very different transformations (see cutting corn and paper in Figure~\ref{fig:samples}). Overall, there is substantial entropy in the correlation statistics illustrating the diversity of our dataset.  
\begin{figure}[t]
\begin{center}
  \includegraphics[width=1.0\columnwidth]{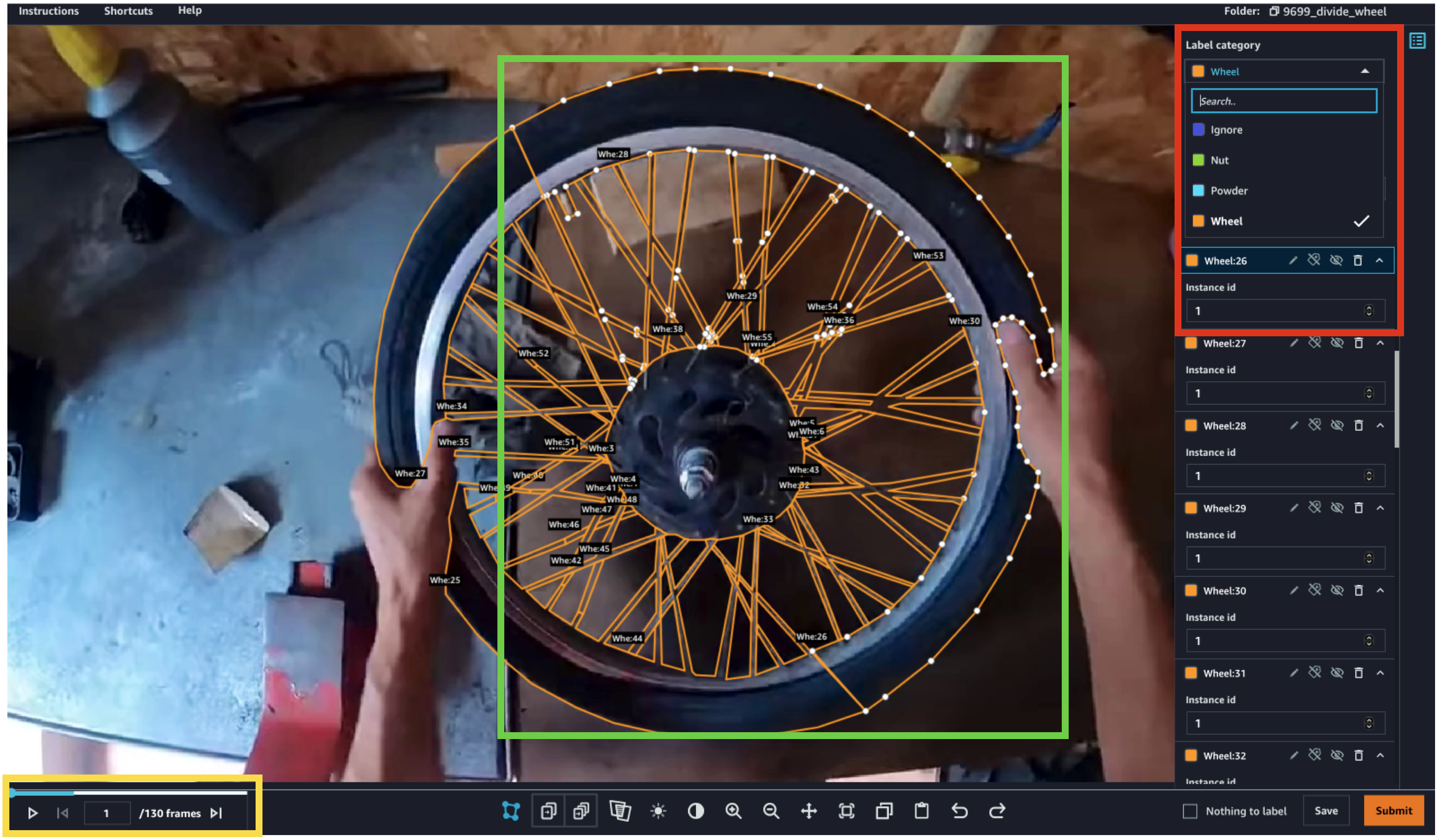}
\end{center}
\vspace{-3mm}
  \caption{
   Interface of our annotation tool. Objects are annotated with polygons (shown in green), and additional “Category” and “Instance id” labels (red). Annotations are automatically propagated to the next frame and then manually adjusted (yellow). 
  }
\vspace{-3mm}
\label{fig:inter}
\end{figure}

\subsection{Annotation collection}
To label the videos selected above, we begin by adjusting the temporal boundaries of each clip to tightly enclose the entire duration of the transformation, with the exception of extremely long sequences (a minute or longer). To balance the cost and temporal density of the annotations we choose to label videos at 5 fps.  

A key question is how to annotate objects as they split into parts (e.g. due to cutting or breaking). To avoid ambiguity, we adopt the most straightforward and general principle: if a region is marked as an object in the first frame of a video, all the parts that originate from it maintain the same identity. For example, the yolks from the broken eggs in Figure~\ref{fig:samples} maintain the identity of the object they originated from. This approach also ensures that there is an unambiguous signal in the data (spatio-temporal continuity) that algorithms can use to achieve generalization.

There are, however, examples in which it is impossible to provide an accurate instance mask for a region. In the second row of Figure~\ref{fig:samples} we show two such cases. In the first one, a piece of clay is experiencing fast motion, making establishing a clear boundary impossible. In the second example, the egg whites from several eggs are mixed together, making it impossible to separate them from each other. Rather than skipping such videos, we choose to label the ambiguous regions with tight ``Ignore'' segments (shown in gray in the figure), which are not used at either training or evaluation time. This flexible approach allows us to consistently annotate even the most challenging videos.

Given the complexity of the task, we hired a fixed team of 20 professional annotators for the entire duration of the project. They received detailed instructions on the task and edge cases which we detail in Section~\ref{sec:instr}. The annotators were first trained for 4 weeks to ensure consistent behavior. Each video was labeled by one annotator using Amazon SageMaker GroundTruth tool for polygon labeling shown in Figure~\ref{fig:inter}. For videos featuring multiple objects and an additional “Instance id” label was provided. The videos were then reviewed by a small, held-out group of skilled annotators and returned to the original worker for correction. This process was repeated until no more issues could be identified. On average, 3.9 annotation-review cycles were performed for each video to ensure the highest label quality.

Overall 175,913 masks were collected, with an average track duration of 21.3 seconds. We report additional statistics of the dataset in Section~\ref{sec:stat} in the appendix.

\subsection{Splits and metrics}
VOST is split into 572 train, 70 validation, and 71 test videos. We have released the labels for train and validation sets, but the test set is held out and only accessible via an evaluation server to prevent over-fitting. Furthermore, we ensure that all three sets are well separated by enforcing that each kitchen from~\cite{damen2022rescaling} and each subject from~\cite{grauman2022ego4d} appears in only one of the train, validation or test sets.
\definecolor{Gray}{gray}{0.85}
\newcolumntype{g}{>{\columncolor{Gray}}C}
\begin{table*}[bt]
\centering
\small
\begin{tabularx}{\linewidth}{l  C{0.115\linewidth} C{0.115\linewidth} C{0.115\linewidth} C{0.115\linewidth} g{0.115\linewidth} g{0.115\linewidth}}
\toprule
 &   \multicolumn{2}{c}{VOST val} & \multicolumn{2}{c}{VOST test} & \multicolumn{2}{c}{DAVIS'17 val}  \\ 
\cmidrule(lr){2-3} \cmidrule(lr){4-5} \cmidrule(lr){6-7}
& $\mathcal{J}_{tr}$ & $\mathcal{J}$ & $\mathcal{J}_{tr}$   &  $\mathcal{J}$ & $\mathcal{J}_{tr}$ & $\mathcal{J}$     \\
\cmidrule{1-7}
OSMN Match~\cite{Yang2018osmn}    & 7.0 & 8.7 & 8.5 & 10.2 & ~41.3 &  ~49.6    \\
\rowcolor{lavender} OSMN Tune~\cite{Yang2018osmn}    & 17.6 & 23.0  & 20.1 & 26.1 &\cellcolor{Gray} 57.2 &\cellcolor{Gray} 68.3    \\
CRW~\cite{jabri2020space} & 13.9  & 23.7  & 20.8 & 28.0 & ~53.6 & ~64.4  \\
\rowcolor{lavender} CFBI\cite{yang2020collaborative} & 32.0 & 45.0  & 32.1 & 43.9 &\cellcolor{Gray} 75.0 &\cellcolor{Gray} 79.3    \\
CFBI+\cite{yang2021collaborative} & 32.6 & 46.0 & 31.6 &  46.7 & ~76.3 & ~80.1  \\
\rowcolor{lavender} AOT\cite{yang2021associating} & \textbf{36.4}  & \textbf{48.7}  & \textbf{37.1} & \textbf{49.9} &\cellcolor{Gray} 80.4 &\cellcolor{Gray} 82.3    \\
XMem\cite{cheng2022xmem} & 33.8 & 44.1  &  32.0 & 44.0 & ~\textbf{81.1} & ~\textbf{82.9}    \\
\rowcolor{lavender} HODOR Img\cite{athar2022hodor} & 13.9 & 24.2  & 22.1 & 29.0 &\cellcolor{Gray} 70.2 &\cellcolor{Gray} 74.7    \\
HODOR Vid\cite{athar2022hodor} & 25.4  & 37.1  & 27.6 & 42.0 & ~74.0 & ~77.4    \\
\bottomrule
\end{tabularx}
\caption{Benchmarking existing methods on VOST. We report results on both validation and test sets of our dataset, using IoU after transformation $\mathcal{J}_{tr}$ as well as the overall IoU $\mathcal{J}$. We include DAVIS'17 val scores for reference. Performance of all methods is 2.2-5.9 times lower in terms of $\mathcal{J}_{tr}$ on VOST compared to DAVIS, emphasizing the complexity of the problem.} \vspace{1mm}
\label{tab:sota}
\vspace{-5mm}
\end{table*}

For evaluation, traditionally, video object segmentation datasets use a combination of region similarity $\mathcal{J}$ and contour accuracy $\mathcal{F}$~\cite{pont20172017,xu2018youtube}. The former is the standard intersection-over-union~\cite{everingham2010pascal} between the predicted $M$ and ground truth masks $G$, which captures the fraction of pixels that are correctly labeled. Contour accuracy, on the other hand, measures how accurate the boundaries of the predicted masks are~\cite{martin2004learning}. Both quantities are computed separately for each instance in each frame and then averaged over frames in a video and over instances. 

We propose two modifications to the standard metrics to better reflect our problem setting. Firstly, we note that contours are often not well defined for the kind of masks we are dealing with: some objects are semi-transparent, and the amount of motion blur is significant. Thus, we do not measure contour accuracy in our experiments. Secondly, recall that region similarity $\mathcal{J}$ for every object $o_i$ is averaged over all video frames:
\begin{equation}
\mathcal{J}(o_i, F)=\frac{1}{|F|} \sum_{f \in F}\mathcal{J}(M^f_{o_i}, G^f_{o_i}),
\end{equation}
where $F$ is the set of frames and $M^f_{o_i}, G^f_{o_i}$ are the predicted and ground truth masks for object $o_i$ in frame $f$ respectively. Hence, every frame has an equal influence on the overall score. This is adequate for the standard VOS setting, but we are interested not in how well a method can segment an object overall, but in how robust it is to transformations. To reflect this fact, we separately measure the region similarity after the transformation has been mostly completed: $\mathcal{J}_{tr} = \mathcal{J}(o_i, \hat{F})$,
where $\hat{F}$ represents the last 25\% of the frames in a sequence. We report both $\mathcal{J}$ and $\mathcal{J}_{tr}$ in our experiments, but use the latter as the main metric.

\section{Analysis of the State-of-the-art Methods}
\label{sec:anal}
We now use VOST to analyze how well can existing VOS methods handle object transformations. All the models are initialized from their best DAVIS'17 checkpoint (usually pre-trained on large-scale image and/or video collection) and fine-tuned on the training set of VOST, unless stated otherwise. We use the original implementations, only adapting the loss to correctly handle ``Ignore'' labels and tuning the number of training iterations on the validation set. More details are provided in Section~\ref{sec:impl} in the appendix.

\subsection{Methods}
We evaluate a total of nine video segmentation algorithms and their variants, which are selected to cover the main trends in the field over recent years. In addition, the methods' performance on existing benchmarks and public availability of the code were taken into account. 

We include OSMN~\cite{Yang2018osmn} as a representative approach for early deep-learning methods that either fine-tune a CNN on the first frame (denoted as OSMN Tune) or employ a more efficient matching mechanism (OSMN Match). As a complementary approach, we evaluate the self-supervised CRW objective~\cite{jabri2020space} for mask propagation which only uses local information between consecutive frame pairs.

In the mainstream, memory-based family of methods we evaluate CFBI~\cite{yang2020collaborative} and its improved variant CFBI+~\cite{yang2021collaborative}, which have been established as very strong baselines on existing benchmarks. In addition, we include the transformer-based AOT approach~\cite{yang2021associating}, and the very recent XMem framework~\cite{cheng2022xmem}, which specifically focuses on long videos.

Finally, we study another recent method - HODOR~\cite{athar2022hodor}, which performs template matching on the object, not on the patch level. We include both the image-based version of this approach, which is trained on COCO (denoted as HODOR Img), as well as the video-based one (HODOR Vid). 

\subsection{Results}
\smallsec{Can existing methods handle transformations?} In Table~\ref{tab:sota} we start by reporting the performance of approaches described above on the validation and test sets of VOST. For reference, we also report the performance of these methods on the validation set of DAVIS'17 on the right. 

Firstly, we observe that the appearance matching baseline (OSMN Match in the table) fails dramatically. This is to be expected as virtually all videos in our dataset feature major appearance changes. Expensive test time fine-tuning on the first frame of a video (OSMN Tune) improves the performance of this baseline, but the validation set score remains 3.3 times lower than on DAVIS. Local mask propagation used by CRW is more robust to appearance change, but cannot handle occlusions, which are plentiful in first-person videos, and hence also struggles on VOST.   

Next, we see that the more advanced, memory-based methods (rows 4 to 7 in the table) are indeed more capable due to their efficient mechanism for updating the appearance model of the target. That said, performance remains low, with the gap between $\mathcal{J}_{tr}$ and $\mathcal{J}$ on VOST being especially large. On DAVIS, on the other hand, the gap is almost completely eliminated by the most recent AOT and XMem baselines. These results demonstrate that, while memory-based methods are capable of segmenting objects through minor appearance changes caused by translations and deformations, they fail under more challenging transformations.

Another notable observation is that the image-based HODOR baseline (HODOR Img in the table), which is only trained on COCO, shows a major loss in performance compared to DAVIS. This illustrates that static object models learned from images break when objects start to transform. Moreover, the variant of this model trained on videos also underperforms, indicating that object-level matching might not be the optimal approach when object shape and appearance change significantly during the video.

\smallsec{What makes the problem challenging?} Significant change in object shape and appearance is one factor that is common to virtually all the videos in VOST. We now analyze a representative subset of the baselines more closely to identify their additional failure modes. To this end, in Table~\ref{tab:abl} we report the $\mathcal{J}_{tr}$ score on subsets of the validation set characterized by various \textit{quantifiable} challenges, such as the length of the video, or presence of occlusions. 
\begin{table}[bt]
 \centering
  {
\resizebox{\columnwidth}{!}{
    \begin{tabular}{l@{\hspace{1em}}|c@{\hspace{1em}}c@{\hspace{1em}}c@{\hspace{1em}}c@{\hspace{1em}}}
      & OSMN Tune~\cite{Yang2018osmn}  &
      CFBI+\cite{yang2021collaborative} & AOT~\cite{yang2021associating} & HODOR Vid~\cite{athar2022hodor}                  \\\hline
     All & 17.6 (-0.0) & 32.6 (-0.0) & 36.4 (-0.0) & 25.4 (-0.0)            \\
     \rowcolor{lavender} LNG & 12.4 (-5.2) & 30.4 (-2.2) & 34.7 (-1.7) & 25.0 (-0.4)          \\
     MI & 14.7 (-2.9) & 26.4 (-6.2) & 27.2 (-9.2) & 20.6 (-4.8)           \\
     \rowcolor{lavender} OCC & 17.2 (-0.4) & 28.1 (-4.5) & 30.7 (-5.7) & 17.6 (-7.8)           \\
     FM & 17.0 (-0.6) & ~21.8 (-10.7) & ~23.8 (-12.5) & 16.0 (-9.4)          \\
     \rowcolor{lavender} SM & 14.4 (-3.2) & 23.3 (-9.2) & ~24.7 (-11.7) & 16.6 (-8.8)       \\
\end{tabular}
}
}
\caption{Quantitative evaluation of failure modes of a subset of the baselines on the validation set using $\mathcal{J}_{tr}$. We analyze such factors as video length (LNG), presence of several instances (MI), occlusions (OCC), fast object motion (FM) and small objects (SM).}
\label{tab:abl}
\vspace{-5mm}
\end{table}

Firstly, by evaluating on videos that are longer than 20 seconds (indicated with LNG in the table), we observe that length alone does not present a significant challenge for most of the methods. This demonstrates that the complexity of the problem is associated with the content of our dataset (object transformations), not with the technical challenges of processing long sequences. 

One unique aspect of our task is that the objects in multi-instances sequences are typically close in appearance (e.g. several eggs). Evaluating on such sequences (indicated with MI in the table) significantly reduces the performance of all the methods. It is not surprising, as appearance-first models are especially ill-suited for this scenario. Intriguingly, the object-level matching strategy of HODOR Vid as well as the expensive test time fine-tuning of OSMN Tune, although less effective overall, seems to be more robust to multi-instance segmentation.

Next, we look at two aspects that test the methods' object permanence capabilities  - full occlusions (denoted as OCC in the table) and fast motion (FM), which is often associated with objects going out of frame. The latter is measured as the distance between object centers in consecutive frames normalized by the object size (see Section~\ref{sec:stat} for details). Interestingly, the simplest OSMN Tune baseline is the most robust to object disappearance. More advanced methods rely heavily on the objects being visible throughout the video and struggle in highly dynamic scenes. 

Finally, our dataset features many small objects (denoted as SM in the table), which are equally challenging for all methods. Overall, we can conclude that reliance on appearance cues and the lack of spatio-temporal modeling capabilities (e.g. modeling object permanence) are some of the main limitations of existing approaches.

\smallsec{Are these challenges easy to address?} After we have observed that VOST features many challenges that are underrepresented in existing benchmarks, it is natural to ask if we can modify the top-performing AOT baseline to address at least some of them. To this end, in Table~\ref{tab:arch} we explore several intuitive directions. Firstly, we increase the length of the training sequences from 5 to 15 frames. While this leads to some improvements, they are limited as the model is ill-equipped to capitalize on longer-term temporal cues.

Next, we increase the spatio-temporal modeling capacity of AOT by replacing the short-term memory module, which uses a transformer to match the patches in the current and previous frame, with a recurrent transformer (denoted as R-STM in the table, details are provided in Section~\ref{sec:rstm} in the appendix). It is more similar to classical recurrent architectures like ConvGRU~\cite{ballas2015delving} and can aggregate a rich spatio-temporal representation of a video over time. This modification translates to stronger transformation modeling capabilities, as indicated by the improved $\mathcal{J}_{tr}$ score.
\begin{table}[bt]
 \centering
  {
\resizebox{\columnwidth}{!}{
    \begin{tabular}{l@{\hspace{1em}}|c@{\hspace{1em}}c@{\hspace{1em}}c@{\hspace{1em}}c@{\hspace{1em}}c@{\hspace{1em}}}
      & AOT~\cite{yang2021associating}  & + 15 fr. & + R-STM & + 10 fps. &   + m-s.    \\\hline
     $\mathcal{J}_{tr}$ & 36.4 & 37.4 & 38.5 & 40.7 &  40.1     \\
     $\mathcal{J}$ & 48.7  & 49.2  & 49.7   & 51.9   & 52.3        \\
 
\end{tabular}
}
}
\caption{Addressing some of the limitations of AOT. We experiment with training on longer sequences, replacing short-term memory with a recurrent transformer and increasing temporal and spatial resolution.}
\label{tab:arch}
\vspace{-3mm}
\end{table}
\begin{figure}[t]
\begin{center}
  \includegraphics[width=0.9\columnwidth]{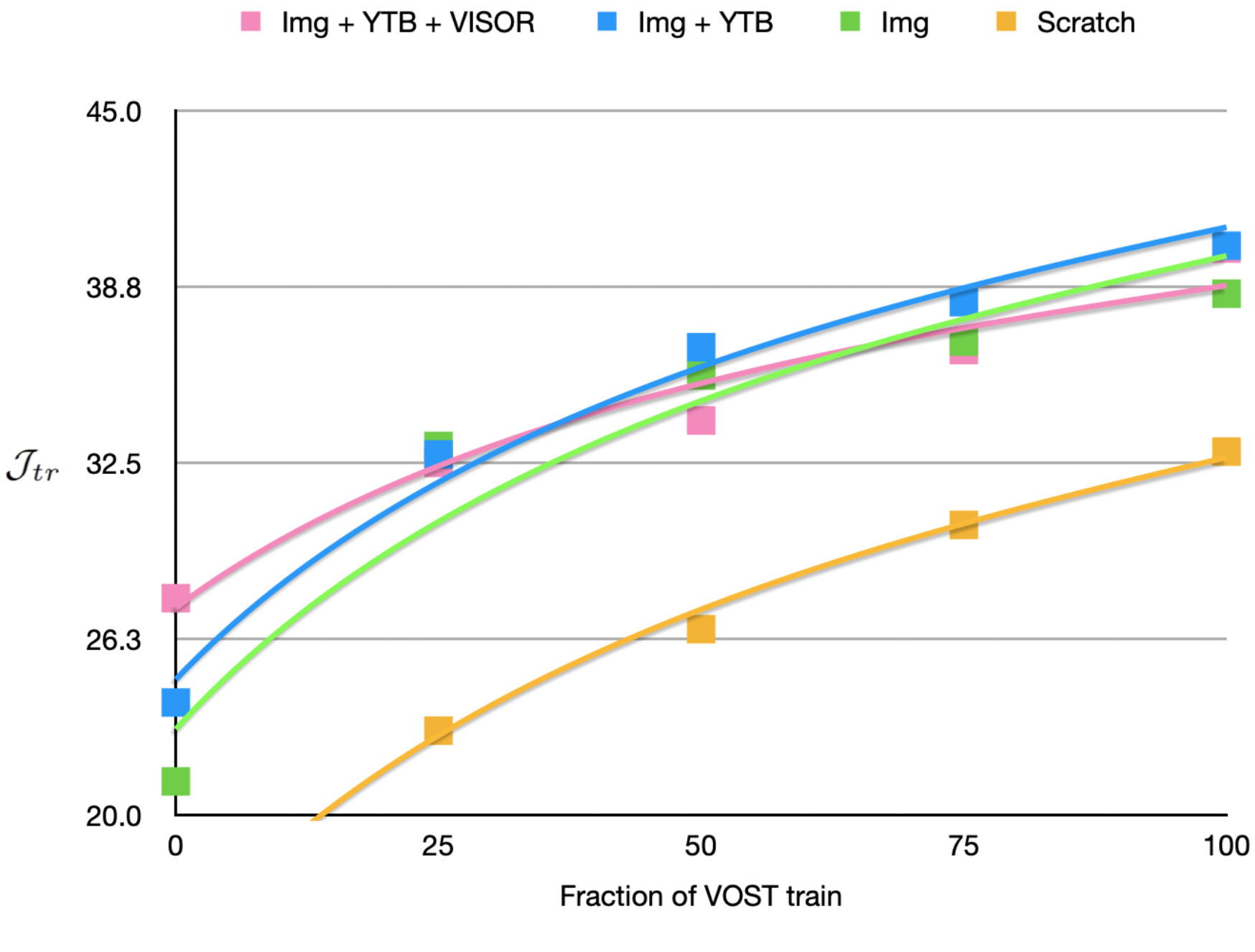}
\end{center}
\vspace{-5mm}
  \caption{
   Evaluation of the effect of the training set size on AOT+ using $\mathcal{J}_{tr}$ on the validation set of VOST. We investigate both the effect of pre-training (on static images and videos) and the fraction of VOST train used for fine-tuning. 
  }
\vspace{-5mm}
\label{fig:data}
\end{figure}

\begin{figure*}[t]
    \centering
    \includegraphics[width=1\linewidth]{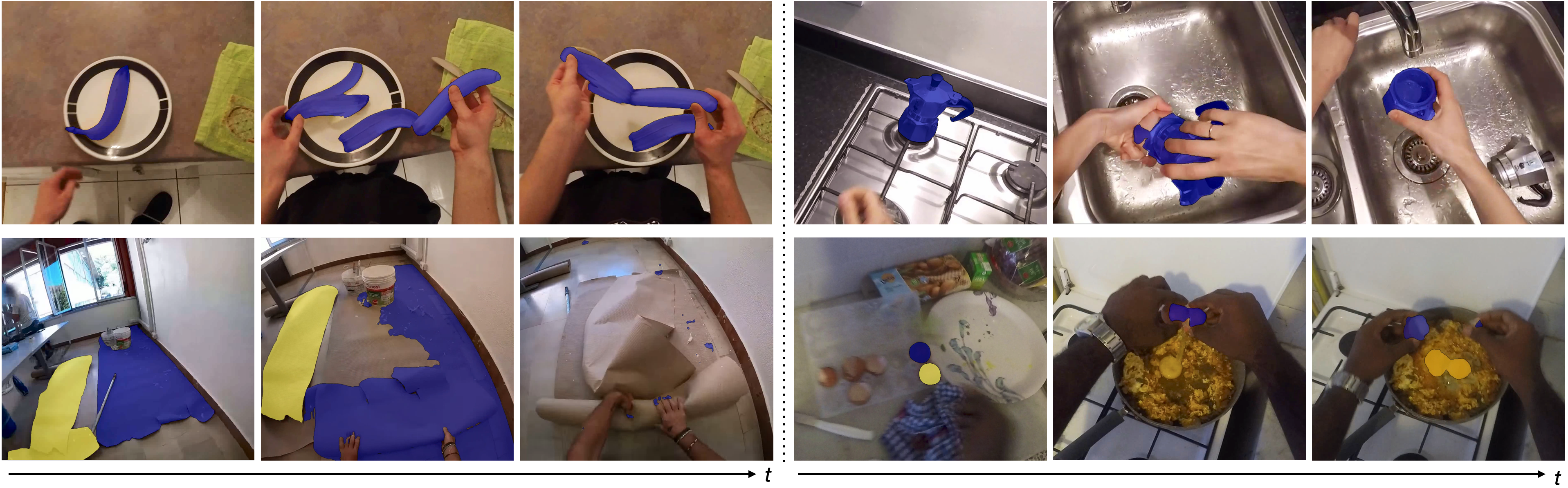}
    \vspace{-6mm}
    \caption{Qualitative results of our AOT+ baseline on sequences from validation and test sets of VOST (see \href{https://youtu.be/SBdA6HCXf_M}{video} for full results). Colours represent instance ids. We can see that, while existing, appearance-first methods can handle relatively challenging transformations, they struggle in the most testing scenarios when appearance is either not enough to distinguish between objects or it changes dramatically. }
    \label{fig:qual}
    \vspace{-3mm}
\end{figure*}
Finally, we experiment with increasing the temporal and spatial resolution of the model at test time by evaluating at 10 fps and enabling multi-scale inference (denoted as m-s. in the table). Both modifications increase the overall performance $\mathcal{J}$, but, notably, the spatial resolution has a smaller effect and even decreases $\mathcal{J}_{tr}$ somewhat. This result suggests that accurately modeling fine-grained temporal information is key for achieving progress on VOST. 

We qualitatively analyze both the success and failure modes of the final variant, which we denote as AOT+, in Figure~\ref{fig:qual}. Firstly, we can see that this model can perfectly handle the banana peeling sequence, illustrating its robustness to relatively challenging transformations. However, in the next sequence the limitations of appearance-first approaches start to show. AOT+ first confuses the coffeemaker with the hand due to reflection and then fails to separate the top part of the objects from the metal sink. 
Next, in the paper rolling sequence, the two instances are correctly segmented at first, but as soon as they are moved and folded together the track of identities is lost and the model breaks down. Finally, AOT+ fails completely in the very challenging egg-cracking example, being not able to both segment the full extents of the eggs and to distinguish between them. 

\smallsec{Is more data all you need?} We now investigate whether the challenges we saw above can be addressed by simply training a model like AOT+ on a larger dataset. To this end, we vary both the pre-training datasets and the size of the VOST training set itself and report the results in Figure~\ref{fig:data}. 

Firstly, we see that although pre-training is important the static image dataset proposed in~\cite{liang2020video} is enough to provide a strong initialization and further pre-training on videos from YTVOS only brings marginal improvements. We additionally experiment with pre-training on the very recent VISOR dataset~\cite{darkhalil2022epic}, which is also sourced from EPIC-KITCHENS, but features only few transformations. In-domain pre-training indeed improves zero-shot performance but does not bring noticeable benefits after fine-tuning on VOST. 

Finally, we observe that, while increasing the size of the training set of VOST does have a noticeable effect on performance, the improvements quickly saturate. If we extrapolate the trend, it would require labeling at least 30,000 videos with complex object transformations for AOT+ to reach the score of 80.0 on $\mathcal{J}_{tr}$, which is not practical. 

\section{Discussion and Limitations}
\label{sec:disc}
In this work, we demonstrated that segmenting objects through transformations presents novel challenges, which existing algorithms are ill-equipped to address. Our analysis provides insights into the failure modes of these methods, while further raising a number of important questions. 

\smallsec{Ambiguity} is inevitable when dealing with object transformations. When designing VOST, we have put a lot of effort to make the annotations as consistent as possible. To this end, we followed the established opinion in cognitive science literature~\cite{spelke1990principles,kahneman1992reviewing} that object perception is driven by universal principles, such as spatio-temporal cohesion and object permanence. We further ensured that for a few scenarios that cannot be resolved on the basis of these principles alone the ``Ignore'' label is used. That said, providing additional annotations, for example, in the form of semantic labels for objects parts, could further enrich the dataset.

\smallsec{Data} plays a key role in deep learning, however, our analysis in Figure~\ref{fig:data} demonstrates that pre-training on generic video collections does not result in significant improvements on VOST. What is needed is large amounts of data featuring object transformations. As we have shown in Section~\ref{sec:select}, collecting such videos requires a lot of effort. Automatic data collection using recent, self-supervised visual-language models~\cite{singh2022flava,alayrac2022flamingo} is a promising way to scale the dataset and extend it to third-person videos.

\smallsec{Model architectures} are another important dimension of the problem. The capacity of a model is what determines how effectively it can use the available data. In Table~\ref{tab:arch} we have shown that, while extending the spatio-temporal capabilities of existing approaches can help, incremental improvements do not address the most fundamental challenges. An entirely new approach to modeling objects in videos is needed, with the recent spatio-temporal transformer architectures~\cite{bertasius2021,arnab2021vivit} being a possible candidate.

{\small
\bibliographystyle{ieee_fullname}
\bibliography{egbib}

\begin{thebibliography}{10}\itemsep=-1pt

\bibitem{alayrac2022flamingo}
Jean-Baptiste Alayrac, Jeff Donahue, Pauline Luc, Antoine Miech, Iain Barr,
  Yana Hasson, Karel Lenc, Arthur Mensch, Katie Millican, Malcolm Reynolds,
  et~al.
\newblock Flamingo: a visual language model for few-shot learning.
\newblock In {\em NeurIPS}, 2022.

\bibitem{arnab2021vivit}
Anurag Arnab, Mostafa Dehghani, Georg Heigold, Chen Sun, Mario Lu{\v{c}}i{\'c},
  and Cordelia Schmid.
\newblock {ViVIT}: A video vision transformer.
\newblock In {\em ICCV}, 2021.

\bibitem{athar2022hodor}
Ali Athar, Jonathon Luiten, Alexander Hermans, Deva Ramanan, and Bastian Leibe.
\newblock {HODOR}: High-level object descriptors for object re-segmentation in
  video learned from static images.
\newblock In {\em CVPR}, 2022.

\bibitem{avinash2014seamseg}
S Avinash~Ramakanth and R Venkatesh~Babu.
\newblock Seamseg: Video object segmentation using patch seams.
\newblock In {\em CVPR}, 2014.

\bibitem{ba2016layer}
Jimmy~Lei Ba, Jamie~Ryan Kiros, and Geoffrey~E Hinton.
\newblock Layer normalization.
\newblock {\em arXiv preprint arXiv:1607.06450}, 2016.

\bibitem{ballas2015delving}
Nicolas Ballas, Li Yao, Chris Pal, and Aaron Courville.
\newblock Delving deeper into convolutional networks for learning video
  representations.
\newblock In {\em ICLR}, 2016.

\bibitem{bertasius2021}
Gedas Bertasius, Heng Wang, and Lorenzo Torresani.
\newblock Is space-time attention all you need for video understanding?
\newblock In {\em ICML}, 2021.

\bibitem{brox2010object}
Thomas Brox and Jitendra Malik.
\newblock Object segmentation by long term analysis of point trajectories.
\newblock In {\em ECCV}, 2010.

\bibitem{caelles2017one}
Sergi Caelles, Kevis-Kokitsi Maninis, Jordi Pont-Tuset, Laura Leal-Taix{\'e},
  Daniel Cremers, and Luc Van~Gool.
\newblock One-shot video object segmentation.
\newblock In {\em CVPR}, 2017.

\bibitem{chen2020state}
Xi Chen, Zuoxin Li, Ye Yuan, Gang Yu, Jianxin Shen, and Donglian Qi.
\newblock State-aware tracker for real-time video object segmentation.
\newblock In {\em CVPR}, 2020.

\bibitem{chen2018blazingly}
Yuhua Chen, Jordi Pont-Tuset, Alberto Montes, and Luc Van~Gool.
\newblock Blazingly fast video object segmentation with pixel-wise metric
  learning.
\newblock In {\em CVPR}, 2018.

\bibitem{cheng2022xmem}
Ho~Kei Cheng and Alexander~G Schwing.
\newblock {XMem}: Long-term video object segmentation with an atkinson-shiffrin
  memory model.
\newblock In {\em ECCV}, 2022.

\bibitem{cheng2021rethinking}
Ho~Kei Cheng, Yu-Wing Tai, and Chi-Keung Tang.
\newblock Rethinking space-time networks with improved memory coverage for
  efficient video object segmentation.
\newblock In {\em NeurIPS}, 2021.

\bibitem{damen2022rescaling}
Dima Damen, Hazel Doughty, Giovanni~Maria Farinella, Antonino Furnari,
  Evangelos Kazakos, Jian Ma, Davide Moltisanti, Jonathan Munro, Toby Perrett,
  Will Price, et~al.
\newblock Rescaling egocentric vision: collection, pipeline and challenges for
  {EPCI-KITCHENS-100}.
\newblock {\em International Journal of Computer Vision}, 130(1):33--55, 2022.

\bibitem{darkhalil2022epic}
Ahmad Darkhalil, Dandan Shan, Bin Zhu, Jian Ma, Amlan Kar, Richard Ely~Locke
  Higgins, Sanja Fidler, David Fouhey, and Dima Damen.
\newblock {EPIC-KITCHENS VISOR} benchmark: Video segmentations and object
  relations.
\newblock In {\em NeurIPS, Datasets and Benchmarks Track}, 2022.

\bibitem{dave2020tao}
Achal Dave, Tarasha Khurana, Pavel Tokmakov, Cordelia Schmid, and Deva Ramanan.
\newblock {TAO}: A large-scale benchmark for tracking any object.
\newblock In {\em ECCV}, 2020.

\bibitem{devlin2018bert}
Jacob Devlin, Ming-Wei Chang, Kenton Lee, and Kristina Toutanova.
\newblock {BERT}: Pre-training of deep bidirectional transformers for language
  understanding.
\newblock {\em arXiv preprint arXiv:1810.04805}, 2018.

\bibitem{duke2021sstvos}
Brendan Duke, Abdalla Ahmed, Christian Wolf, Parham Aarabi, and Graham~W
  Taylor.
\newblock {SSTVOS}: Sparse spatiotemporal transformers for video object
  segmentation.
\newblock In {\em CVPR}, 2021.

\bibitem{everingham2010pascal}
Mark Everingham, Luc Van~Gool, Christopher~KI Williams, John Winn, and Andrew
  Zisserman.
\newblock The {PASCAL} visual object classes ({VOC}) challenge.
\newblock {\em International journal of computer vision}, 88(2):303--338, 2010.

\bibitem{fan2015jumpcut}
Qingnan Fan, Fan Zhong, Dani Lischinski, Daniel Cohen-Or, and Baoquan Chen.
\newblock {JumpCut}: non-successive mask transfer and interpolation for video
  cutout.
\newblock {\em ACM Trans. Graph.}, 34(6):195--1, 2015.

\bibitem{fillmore1967grammar}
Charles~J Fillmore.
\newblock The grammar of hitting and breaking.
\newblock 1967.

\bibitem{gao2010objects}
Tao Gao and Brian~J Scholl.
\newblock Are objects required for object-files? {R}oles of segmentation and
  spatiotemporal continuity in computing object persistence.
\newblock {\em Visual Cognition}, 18(1):82--109, 2010.

\bibitem{grauman2022ego4d}
Kristen Grauman, Andrew Westbury, Eugene Byrne, Zachary Chavis, Antonino
  Furnari, Rohit Girdhar, Jackson Hamburger, Hao Jiang, Miao Liu, Xingyu Liu,
  et~al.
\newblock {Ego4D}: Around the world in 3,000 hours of egocentric video.
\newblock In {\em CVPR}, 2022.

\bibitem{grundmann2010efficient}
Matthias Grundmann, Vivek Kwatra, Mei Han, and Irfan Essa.
\newblock Efficient hierarchical graph-based video segmentation.
\newblock In {\em CVPR}. IEEE, 2010.

\bibitem{hollingworth2009object}
Andrew Hollingworth and Steven~L Franconeri.
\newblock Object correspondence across brief occlusion is established on the
  basis of both spatiotemporal and surface feature cues.
\newblock {\em Cognition}, 113(2):150--166, 2009.

\bibitem{hu2017maskrnn}
Yuan-Ting Hu, Jia-Bin Huang, and Alexander Schwing.
\newblock Mask-{RNN}: Instance level video object segmentation.
\newblock In {\em NeurIPS}, 2017.

\bibitem{hu2018videomatch}
Yuan-Ting Hu, Jia-Bin Huang, and Alexander~G Schwing.
\newblock Videomatch: Matching based video object segmentation.
\newblock In {\em ECCV}, 2018.

\bibitem{jabri2020space}
Allan Jabri, Andrew Owens, and Alexei Efros.
\newblock Space-time correspondence as a contrastive random walk.
\newblock In {\em NeurIPS}, 2020.

\bibitem{kahneman1992reviewing}
Daniel Kahneman, Anne Treisman, and Brian~J Gibbs.
\newblock The reviewing of object files: Object-specific integration of
  information.
\newblock {\em Cognitive psychology}, 24(2):175--219, 1992.

\bibitem{khoreva2019lucid}
Anna Khoreva, Rodrigo Benenson, Eddy Ilg, Thomas Brox, and Bernt Schiele.
\newblock Lucid data dreaming for video object segmentation.
\newblock {\em International Journal of Computer Vision}, 127(9):1175--1197,
  2019.

\bibitem{levin1993english}
Beth Levin.
\newblock {\em English verb classes and alternations: A preliminary
  investigation}.
\newblock University of Chicago press, 1993.

\bibitem{li2013video}
Fuxin Li, Taeyoung Kim, Ahmad Humayun, David Tsai, and James~M Rehg.
\newblock Video segmentation by tracking many figure-ground segments.
\newblock In {\em ICCV}, 2013.

\bibitem{li2020delving}
Yuxi Li, Ning Xu, Jinlong Peng, John See, and Weiyao Lin.
\newblock Delving into the cyclic mechanism in semi-supervised video object
  segmentation.
\newblock {\em NeurIPS}, 2020.

\bibitem{liang2021video}
Shuxian Liang, Xu Shen, Jianqiang Huang, and Xian-Sheng Hua.
\newblock Video object segmentation with dynamic memory networks and adaptive
  object alignment.
\newblock In {\em ICCV}, pages 8065--8074, 2021.

\bibitem{liang2020video}
Yongqing Liang, Xin Li, Navid Jafari, and Jim Chen.
\newblock Video object segmentation with adaptive feature bank and
  uncertain-region refinement.
\newblock {\em NeurIPS}, 2020.

\bibitem{lin2014microsoft}
Tsung-Yi Lin, Michael Maire, Serge Belongie, James Hays, Pietro Perona, Deva
  Ramanan, Piotr Doll{\'a}r, and C~Lawrence Zitnick.
\newblock Microsoft {COCO}: Common objects in context.
\newblock In {\em ECCV}, 2014.

\bibitem{luiten2018premvos}
Jonathon Luiten, Paul Voigtlaender, and Bastian Leibe.
\newblock Premvos: Proposal-generation, refinement and merging for video object
  segmentation.
\newblock In {\em ACCV}, 2018.

\bibitem{martin2004learning}
David~R Martin, Charless~C Fowlkes, and Jitendra Malik.
\newblock Learning to detect natural image boundaries using local brightness,
  color, and texture cues.
\newblock {\em IEEE Transactions on Pattern Analysis and Machine Intelligence},
  26(5):530--549, 2004.

\bibitem{oh2018fast}
Seoung~Wug Oh, Joon-Young Lee, Kalyan Sunkavalli, and Seon~Joo Kim.
\newblock Fast video object segmentation by reference-guided mask propagation.
\newblock In {\em CVPR}, 2018.

\bibitem{oh2019video}
Seoung~Wug Oh, Joon-Young Lee, Ning Xu, and Seon~Joo Kim.
\newblock Video object segmentation using space-time memory networks.
\newblock In {\em ICCV}, 2019.

\bibitem{perazzi2017learning}
Federico Perazzi, Anna Khoreva, Rodrigo Benenson, Bernt Schiele, and Alexander
  Sorkine-Hornung.
\newblock Learning video object segmentation from static images.
\newblock In {\em CVPR}, 2017.

\bibitem{perazzi2016benchmark}
Federico Perazzi, Jordi Pont-Tuset, Brian McWilliams, Luc Van~Gool, Markus
  Gross, and Alexander Sorkine-Hornung.
\newblock A benchmark dataset and evaluation methodology for video object
  segmentation.
\newblock In {\em CVPR}, 2016.

\bibitem{peters2021capturing}
Benjamin Peters and Nikolaus Kriegeskorte.
\newblock Capturing the objects of vision with neural networks.
\newblock {\em Nature Human Behaviour}, 5(9):1127--1144, 2021.

\bibitem{pont20172017}
Jordi Pont-Tuset, Federico Perazzi, Sergi Caelles, Pablo Arbel{\'a}ez, Alex
  Sorkine-Hornung, and Luc Van~Gool.
\newblock The 2017 {DAVIS} challenge on video object segmentation.
\newblock {\em arXiv preprint arXiv:1704.00675}, 2017.

\bibitem{scholl2007object}
Brian~J Scholl.
\newblock Object persistence in philosophy and psychology.
\newblock {\em Mind \& Language}, 22(5):563--591, 2007.

\bibitem{seong2020kernelized}
Hongje Seong, Junhyuk Hyun, and Euntai Kim.
\newblock Kernelized memory network for video object segmentation.
\newblock In {\em ECCV}, 2020.

\bibitem{seong2021hierarchical}
Hongje Seong, Seoung~Wug Oh, Joon-Young Lee, Seongwon Lee, Suhyeon Lee, and
  Euntai Kim.
\newblock Hierarchical memory matching network for video object segmentation.
\newblock In {\em ICCV}, 2021.

\bibitem{shi2015convolutional}
Xingjian Shi, Zhourong Chen, Hao Wang, Dit-Yan Yeung, Wai-Kin Wong, and
  Wang-chun Woo.
\newblock Convolutional {LSTM} network: A machine learning approach for
  precipitation nowcasting.
\newblock {\em NeurIPS}, 2018.

\bibitem{singh2022flava}
Amanpreet Singh, Ronghang Hu, Vedanuj Goswami, Guillaume Couairon, Wojciech
  Galuba, Marcus Rohrbach, and Douwe Kiela.
\newblock Flava: A foundational language and vision alignment model.
\newblock In {\em CVPR}, 2022.

\bibitem{spelke1990principles}
Elizabeth~S Spelke.
\newblock Principles of object perception.
\newblock {\em Cognitive science}, 14(1):29--56, 1990.

\bibitem{spelke2013perceiving}
Elizabeth~S Spelke.
\newblock Where perceiving ends and thinking begins: The apprehension of
  objects in infancy.
\newblock In {\em Perceptual development in infancy}, pages 209--246.
  Psychology Press, 2013.

\bibitem{spelke2007core}
Elizabeth~S Spelke and Katherine~D Kinzler.
\newblock Core knowledge.
\newblock {\em Developmental science}, 10(1):89--96, 2007.

\bibitem{tokmakov2017learning}
Pavel Tokmakov, Karteek Alahari, and Cordelia Schmid.
\newblock Learning video object segmentation with visual memory.
\newblock In {\em ICCV}, 2017.

\bibitem{tsai2012motion}
David Tsai, Matthew Flagg, Atsushi Nakazawa, and James~M Rehg.
\newblock Motion coherent tracking using multi-label {MRF} optimization.
\newblock {\em International Journal of Computer Vision}, 100(2):190--202,
  2012.

\bibitem{vaswani2017}
Ashish Vaswani, Noam Shazeer, Niki Parmar, Jakob Uszkoreit, Llion Jones,
  Aidan~N Gomez, {\L}ukasz Kaiser, and Illia Polosukhin.
\newblock Attention is all you need.
\newblock In {\em NeurIPS}, 2017.

\bibitem{voigtlaender2019feelvos}
Paul Voigtlaender, Yuning Chai, Florian Schroff, Hartwig Adam, Bastian Leibe,
  and Liang-Chieh Chen.
\newblock {FeelVOS}: Fast end-to-end embedding learning for video object
  segmentation.
\newblock In {\em CVPR}, 2019.

\bibitem{voigtlaender2017online}
Paul Voigtlaender and Bastian Leibe.
\newblock Online adaptation of convolutional neural networks for video object
  segmentation.
\newblock In {\em BMVC}, 2017.

\bibitem{wang2021unidentified}
Weiyao Wang, Matt Feiszli, Heng Wang, and Du Tran.
\newblock Unidentified video objects: A benchmark for dense, open-world
  segmentation.
\newblock In {\em ICCV}, 2021.

\bibitem{wang2019learning}
Xiaolong Wang, Allan Jabri, and Alexei~A Efros.
\newblock Learning correspondence from the cycle-consistency of time.
\newblock In {\em CVPR}, 2019.

\bibitem{xiao2018monet}
Huaxin Xiao, Jiashi Feng, Guosheng Lin, Yu Liu, and Maojun Zhang.
\newblock Monet: Deep motion exploitation for video object segmentation.
\newblock In {\em CVPR}, 2018.

\bibitem{xie2021efficient}
Haozhe Xie, Hongxun Yao, Shangchen Zhou, Shengping Zhang, and Wenxiu Sun.
\newblock Efficient regional memory network for video object segmentation.
\newblock In {\em CVPR}, 2021.

\bibitem{xu2018youtube}
Ning Xu, Linjie Yang, Yuchen Fan, Jianchao Yang, Dingcheng Yue, Yuchen Liang,
  Brian Price, Scott Cohen, and Thomas Huang.
\newblock {YouTube-VOS}: Sequence-to-sequence video object segmentation.
\newblock In {\em ECCV}, 2018.

\bibitem{yang2021self}
Charig Yang, Hala Lamdouar, Erika Lu, Andrew Zisserman, and Weidi Xie.
\newblock Self-supervised video object segmentation by motion grouping.
\newblock In {\em ICCV}, 2021.

\bibitem{Yang2018osmn}
Linjie Yang, Yanran Wang, Xuehan Xiong, Jianchao Yang, and Aggelos~K.
  Katsaggelos.
\newblock Efficient video object segmentation via network modulation.
\newblock {\em CVPR}, 2018.

\bibitem{yang2019unsupervised}
Yanchao Yang, Antonio Loquercio, Davide Scaramuzza, and Stefano Soatto.
\newblock Unsupervised moving object detection via contextual information
  separation.
\newblock In {\em CVPR}, 2019.

\bibitem{yang2020collaborative}
Zongxin Yang, Yunchao Wei, and Yi Yang.
\newblock Collaborative video object segmentation by foreground-background
  integration.
\newblock In {\em ECCV}, 2020.

\bibitem{yang2021associating}
Zongxin Yang, Yunchao Wei, and Yi Yang.
\newblock Associating objects with transformers for video object segmentation.
\newblock In {\em NeurIPS}, 2021.

\bibitem{yang2021collaborative}
Zongxin Yang, Yunchao Wei, and Yi Yang.
\newblock Collaborative video object segmentation by multi-scale
  foreground-background integration.
\newblock {\em IEEE Transactions on Pattern Analysis and Machine Intelligence},
  2021.

\bibitem{zeng2019dmm}
Xiaohui Zeng, Renjie Liao, Li Gu, Yuwen Xiong, Sanja Fidler, and Raquel
  Urtasun.
\newblock {DMM-Net}: Differentiable mask-matching network for video object
  segmentation.
\newblock In {\em ICCV}, 2019.

\end{thebibliography}
}

\clearpage
\appendix

\begin{sidewaysfigure*}[ht]
    \includegraphics[width=1\linewidth]{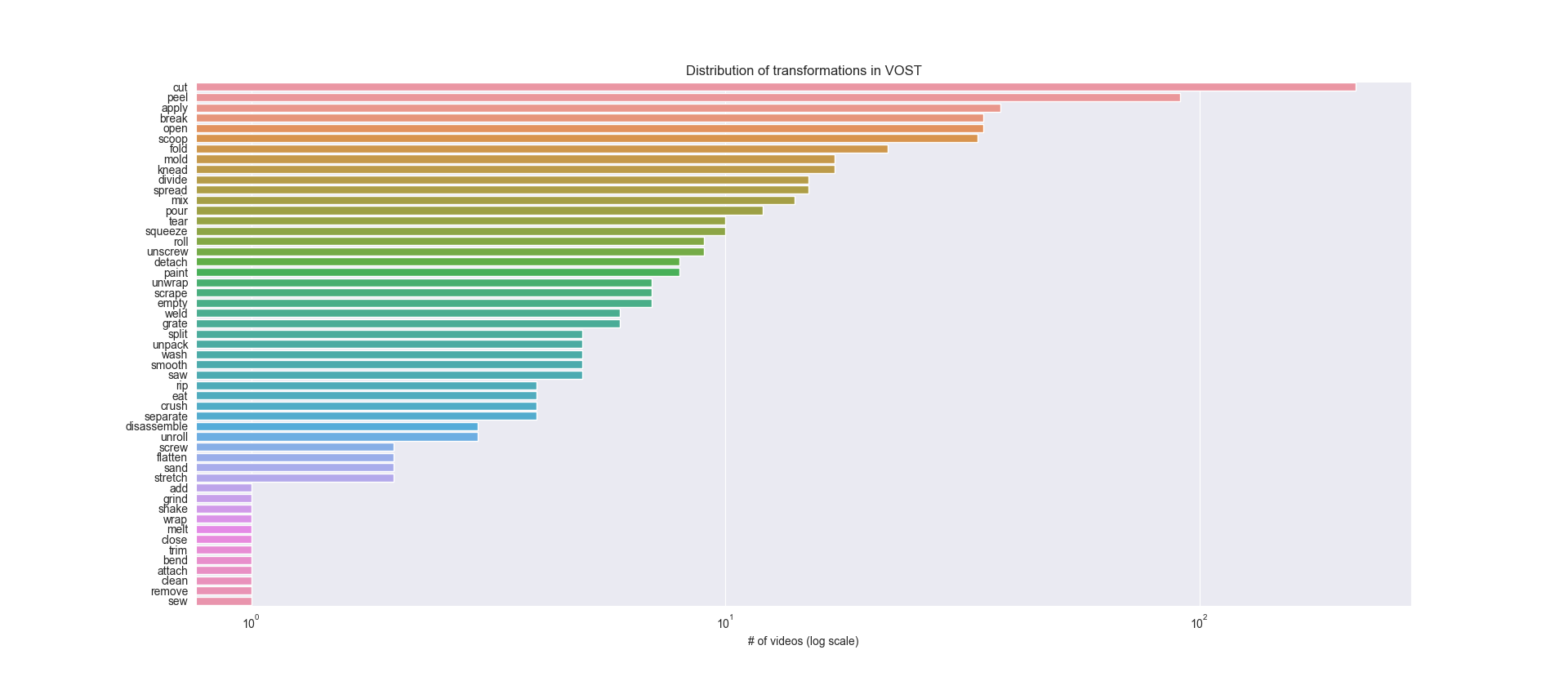}
    \vspace{-5mm}
    \caption{Distribution of transformations in VOST. While there is some bias towards common activities, like cutting and peeling, the tail of the distribution is sufficiently heavy. }
    \label{fig:trans}
\end{sidewaysfigure*}

\begin{figure*}[t]
    \centering
    \includegraphics[width=1\linewidth]{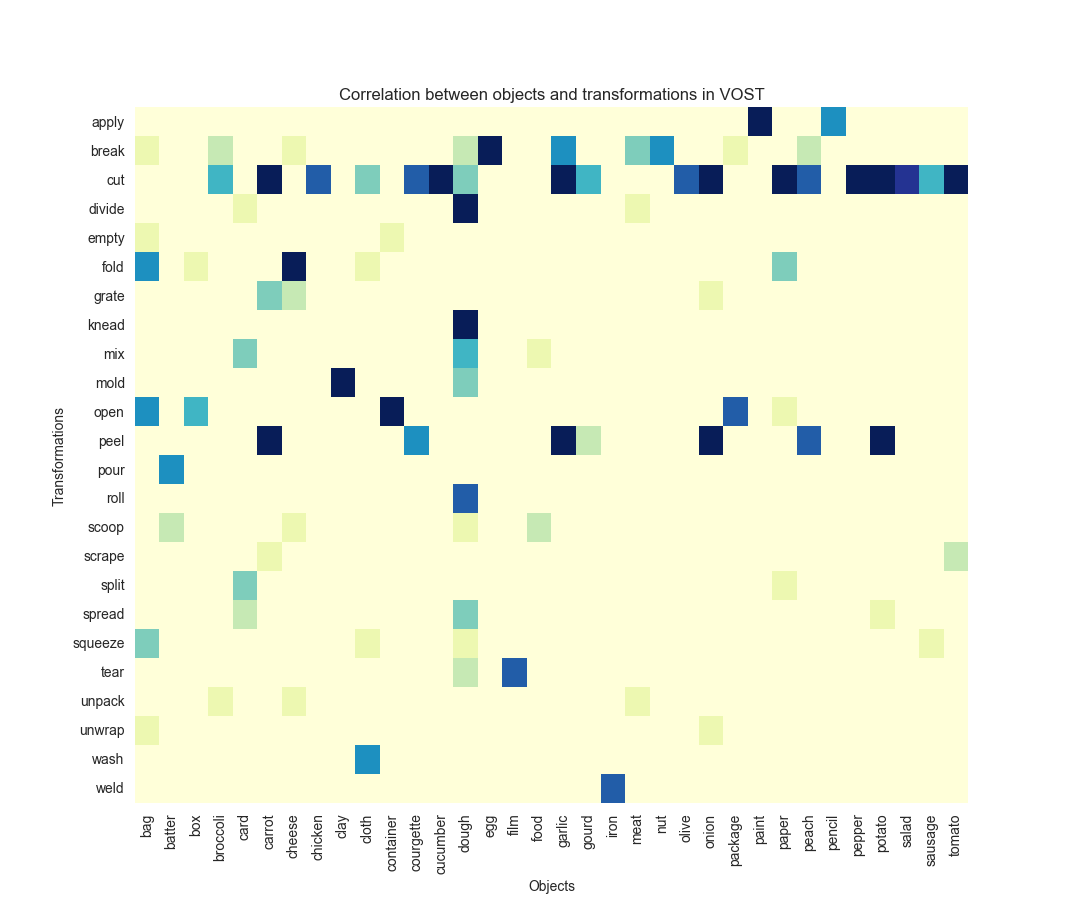}
    \vspace{-5mm}
    \caption{Co-occurrence statistics between the most common transformations and object categories in VOST. We observe that the most common transformation - cutting, has a very broad semantic meaning and can be applied to most objects. Overall, there is substantial entropy in the distribution, illustrating the diversity of VOST.}
    \label{fig:corr}
    \vspace{-3mm}
\end{figure*}

In this appendix, we provide additional visualizations, datasets statistics and implementation details that were not included in the main paper due to space limitations. We begin by describing the contents of the supplementary video, which includes full versions of Figures~\ref{fig:samples} and~\ref{fig:qual} from the main paper in Section~\ref{sec:vid}. We then provide further details on collecting VOST in Section~\ref{sec:coll} and report additional dataset statistics in Section~\ref{sec:stat}. An enlarged version of Figure~\ref{fig:stats} from the main paper is shown in Figures~\ref{fig:trans} and~\ref{fig:corr}. A discussion on limitations of annotation interpolation from~\cite{wang2021unidentified,darkhalil2022epic} in challenging scenarios can be found in Section~\ref{sec:interp}. Finally, we provide the details of our proposed recurrent transformer module in Section~\ref{sec:rstm} and further implementation details in Section~\ref{sec:impl}.

\section{Video Description}
\label{sec:vid}
\subsection{Annotation visualization}
We begin by visualizing VOST annotations for several representative sequences from Figure~\ref{fig:samples} in the main paper in this \href{https://youtu.be/HvNRibYKip0}{video}. Mask colours indicate instance ids, with grey representing ignored regions.

\smallsec{00:00-00:18} In the first sequence we can see 6 separate instance of corn being cut. According to our labeling strategy (see Section~\ref{sec:instr}), only the instances that are being manipulated are labeled in the video. As the objects are separated into many parts and moved around the board, all the parts maintain the identity of the instance they originated from. We can also see that even the smallest parts are labeled with accurate masks. 

\smallsec{00:19-00:54} Next video illustrates the broad semantic meaning of cutting. This sample of paper cutting in the context of making art features many small, thin regions that are accurately labeled, as well as an example of fast motion, when the object is separated into two parts of different colour towards the end of the clip.

\smallsec{00:55-01:15} In this outdoor video a piece of clay is being molded into a brick. In the process, it experiences major shape changes combined with a full occlusion. Moreover, the motions are fast resulting in a significant amount of blur. The corresponding regions are labeled as “Ignore” (shown in gray) to avoid ambiguity during training and evaluation.

\smallsec{01:16-01:40} Finally, the very challenging egg breaking sequence further illustrates our approach to handling ambiguous regions. As the first egg is broken, both the shells as the yolks are labeled with accurate object regions, maintaining the identity of the egg. It is, however, impossible to establish an accurate boundary between the transparent egg white and the bowl, so the annotators label it with a conservative ignore region. As more eggs are broken into the bowl, the yolks are still labeled with correct instance ids but it is impossible to separate the mixed egg whites, so the ignore label is maintained. In addition, this challenging video features fast motion due to objects gong out of frame, but our annotations correctly capture instance ids as the eggs re-appear. 

\subsection{Qualitative results}
We now visualize the outputs of our AOT+ baseline on several sequences from Figure~\ref{fig:qual} in the main paper in the following \href{https://youtu.be/SBdA6HCXf_M}{video}.

\smallsec{00:00-00:19} We begin with a success case where a peeled banana is accurately segmented as it is separated into several parts and its appearance changes. Notice that all the state transitions are smooth in this video and there is a strong contrast between the object and the background, making the task relatively easy for AOT+.

\smallsec{00:20-00:38} In the next sequence, however, although the appearance of the coffeemaker does not change as much, the transitions are more abrupt. Moreover, after the top part is separated and left on the metal sink it experiences full occlusion. This confuses our baseline, which still largely relies on appearance, resulting in the loss of that object part. Moreover, the model experiences several other small failures due to appearance similarity between the coffeemaker and the sink and reflections.

\smallsec{00:39-01:04} Over-reliance on appearance and limited spatio-temporal modeling capabilities of the model cause a complete failure in the next sequence, where two cuts of paper with nearly identical appearance are being rolled together. The model is able to distinguish the objects at first, while they are spatially separated, but as the two instances get mixed together and (self-)occluded it looses track of their identities. 

\smallsec{01:04-01:36} We conclude with the egg breaking example. Here the model has to both deal with major appearance changes and distinguish between the two nearly identical instances. As the first egg is broken it only captures the shell, failing to handle this challenging transformation. AOT+ manages to maintain the eggs identity at first, even though one of them goes out of frame, but ultimately fails at that too when the second egg is broken.

\section{Additional details on Dataset Collection}
\label{sec:coll}
In this section, we first provide the definitions of complexity categories that were used to select VOST videos and report the final distribution of complexity scores. We then report the instructions that were given to the annotators.
\subsection{Complexity categories}
\label{sec:compcat}
Recall that, to focus on challenging object transformations, we labeled all videos from~\cite{damen2022rescaling,grauman2022ego4d} that contained a change of state verb~\cite{fillmore1967grammar,levin1993english} in their original annotation with a complexity score on the scale from 1 to 5. In Table~\ref{tab:comp} we report the definitions of the scores that were used at this stage. Note that the problem of defining what constitutes a complex transformation is inherently ambiguous. The definitions we used are by no means general, but they were helpful to formalize the process of video selection when constructing VOST.
\begin{table}[bt]
 \centering
  {
\resizebox{\columnwidth}{!}{
    \begin{tabular}{l|p{7cm}}
      Score  & Definition \\\hline
     1 & \parbox{7cm}{No visible object transformation. Either the verb was used in a different context or there was a mistake in the original annotation.} \newline \\\hline
     2 & Technically there is a transformation in a video, but it only results in a negligible change of appearance and/or shape (e.g. folding a white towel in half or shaking a paint brush).  \\\hline
     3 & A noticeable transformation that nevertheless preserves the overall appearance and shape of the object (e.g. cutting an onion in half or opening the hood of a car). \\\hline
     4 & A transformation that results in a significant change of the object shape and appearance (e.g. peeling a banana or breaking glass). \\\hline
     5 & Complete change of object appearance, shape and texture (e.g breaking of an egg or grinding beans into flour).  \\
 
\end{tabular}
}
}
\caption{Definition of complexity scores used when filtering videos for VOST. These are by no means general, but they were helpful to formalize the process of video selection when constructing the dataset.}
\label{tab:comp}
\vspace{-2mm}
\end{table}

\begin{figure}[t]
\begin{center}
  \includegraphics[width=\columnwidth]{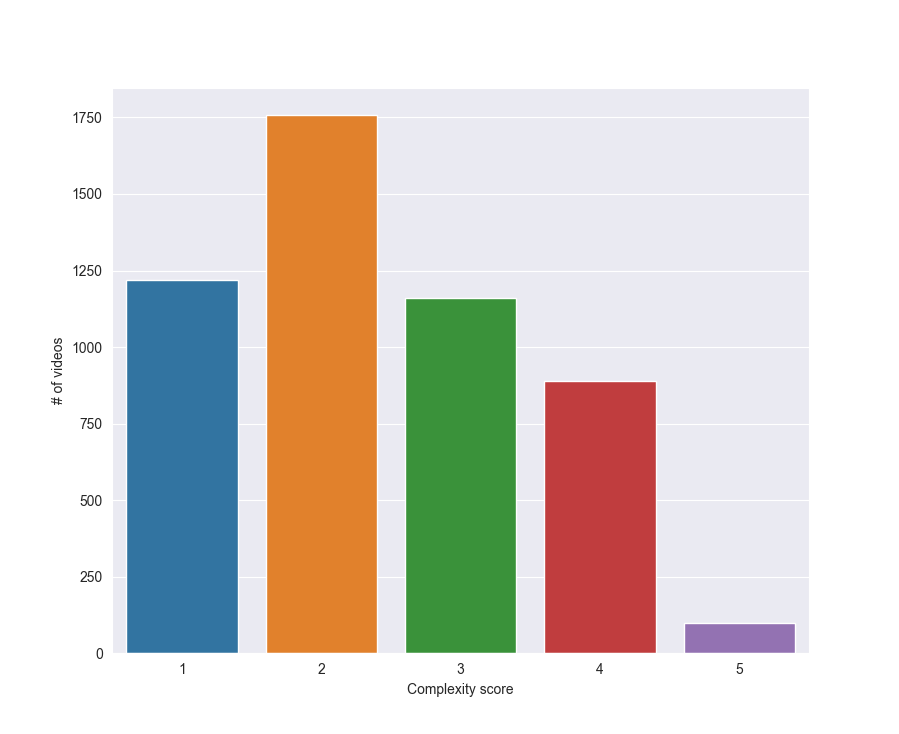}
\end{center}
\vspace{-6mm}
  \caption{
  Distribution of complexity scores among reviewed clips. The majority of the transformations in the wild are not challenging but there is still a sufficient number of clips in the target 4-5 range.
  }
\vspace{-3mm}
\label{fig:comp}
\end{figure}

In Figure~\ref{fig:comp} we report the distribution of complexity score over all the labeled videos. Note that the total number of videos here is significantly lower that the raw number of clips extracted from Ego4D and EPIC-KITCHENS (10,706) because at this stage we also linked the clips representing a consecutive sequence of transformations (e.g. several cuts of the same onion) together. We can see that the wast majority of object transformations in the wild are not challenging, emphasizing the complexity of sourcing videos for VOST. That said, due to the large scale of~\cite{damen2022rescaling,grauman2022ego4d} we are still left with a sufficient number of videos in the target 4-5 range.

\subsection{Annotator instructions}
\label{sec:instr}
We now report the instructions that were used by the annotators to label videos in VOST. The interface of the annotation tool is shown in Figure~\ref{fig:inter} in the main paper.

\begin{itemize}
\item{The goal of this task is to provide polygon annotations for a wide variety of objects as they undergo transformations. The categories of the objects that need to be labeled in each video are provided in the “Label Categories” menu. If there are several objects of a certain category in a video, then only the ones that are being manipulated need to be labeled (e.g. if there are six eggs on a table but only two are broken, then only these 2 should be labeled).}
\item{To label an object, select the appropriate category from the “Label Categories” menu, and then use the polygon tool to draw a polygon around it. A polygon is made up of a series of ordered points that you place around the object. The first and last points of the polygon must be the same and lines (edges) of a polygon cannot cross. When you place the first point, it will turn green. To complete a polygon, close it by selecting the green start point again. }

\item{There is a special label category “Ignore”. It is only to be used in cases when an accurate polygon annotation is impossible to provide for a given region. In particular, there are 4 such scenarios:
\begin{itemize}
\item{Uncertain object boundaries due to motion blur. Label the non-blurred part with a regular polygon, and draw an “Ignore” polygon around the blurry one.}
\item{Tiny object parts that are too small to label to label accurately (e.g. tiny pieces of an onion skin). Draw the smallest possible “Ignore” polygon around each part.}
\item{(Semi)-transparent substances. Treat them in the same way as blurry boundaries (e.g. label the clearly visible part with a regular polygon, and only use the “Ignore” label for the uncertain region).}
\item{Parts of different objects that get mixed to the point at which they cannot be distinguished (e.g. two egg whites getting mixed together).}
\end{itemize}
The “Ignore” label should never be used in the first frame of a video.
}

\item{If an object is (partially) visible through another object (e.g. though a glass bottle), then the corresponding region should be labeled with the category of the front-most object. If that objects is not being labeled in the video, then the “Ignore” label should be used.}

\item{Transformations can result in object splitting (such as breaking a glass). All the parts that results from splitting still need to be labeled (e.g. all the parts of the glass after it has been broken). This includes less obvious examples, such as a bowl wrapped in a plastic foil. As the bowl is getting unwrapped, both the bowl, its content and the plastic wrap need to be annotated. Another example which is worth noting is an egg. As it is getting cracked, both the resulting shells and the egg white/yolk need to be labeled. } 
\item{
If there are multiple objects in a video, use the “Instance id” attribute to indicate which of the polygons belongs to which instance.}
\item{Use the “Copy to next” icon to have the user interface copy all selected polygons (or all polygons if none are selected) in the current frame to the next frame. Use the “Copy to remaining frames” icon to copy all selected or all poylines to all subsequent frames.}
\item{
To adjust the location and shape of a polygon, select the polygon or the label associated with it from the “Labels” list in the menu on the right. Adjust the polygon by moving the points. }
\end{itemize}

\section{Dataset Statistics}
\label{sec:stat}
\begin{figure}[t]
\begin{center}
  \includegraphics[width=\columnwidth]{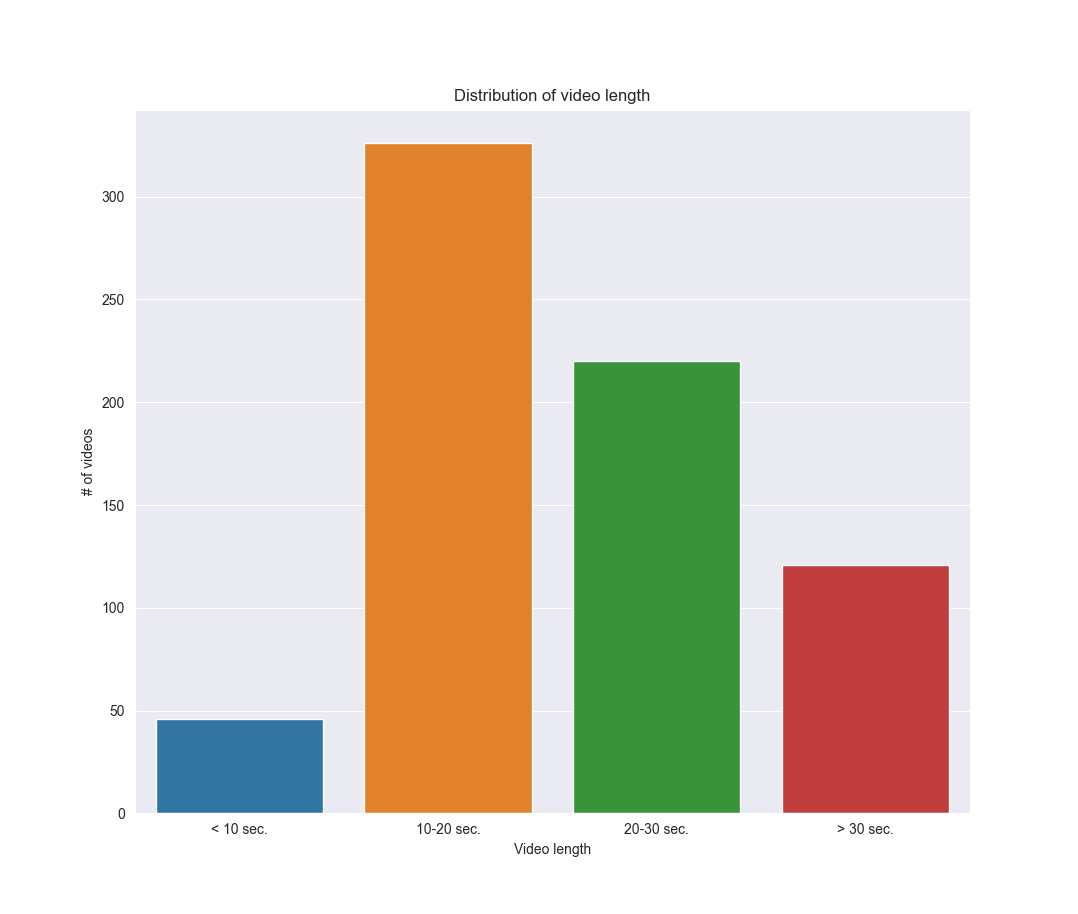}
\end{center}
\vspace{-7mm}
  \caption{
  Distribution of video lengths in VOST. The vast majority of the samples fall into the challenging 10-30 seconds range, and a significant number of the videos are even longer than that.
  }
\vspace{-3mm}
\label{fig:lens}
\end{figure}

In this section, we report additional statistics for the VOST dataset. We begin with Figure~\ref{fig:lens}, which shows the distribution of clip lengths. The wast majority of the videos fall in the challenging 10-30 seconds range, which is significantly longer than in any existing VOS dataset. Moreover, 121 videos are longer than 30 seconds, capturing such lengthy transformations as grinding beans into flour.

\begin{figure}[t]
\begin{center}
  \includegraphics[width=\columnwidth]{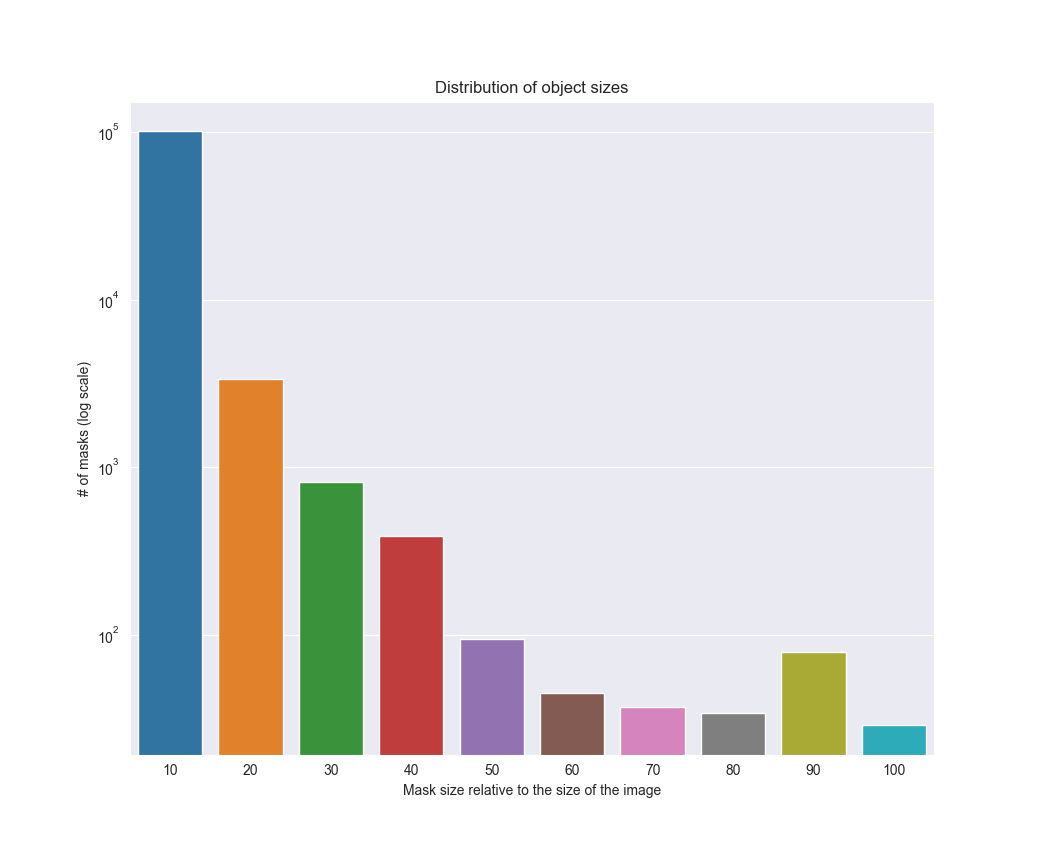}
\end{center}
\vspace{-7mm}
  \caption{
  Distribution of object sizes in VOST. Most of the objects are small due to the nature of first-person videos, but there is a significant long tail of larger objects, such as cars.
  }
\vspace{-3mm}
\label{fig:sz}
\end{figure}

Next, in Figure~\ref{fig:sz} we show the distribution over object mask sizes as a fraction of the whole image. Firstly, we can see that most objects in VOST are small, occupying less than 10\% of the pixels in a frame. This is due to the nature of first-person videos, where the objects that are being manipulated are typically significantly smaller than the person who is manipulating them. That said, the distribution features a significant long tail of larger objects, such as cars or garbage bags, that can occupy more then half of the frame.

\begin{figure}[t]
\begin{center}
  \includegraphics[width=\columnwidth]{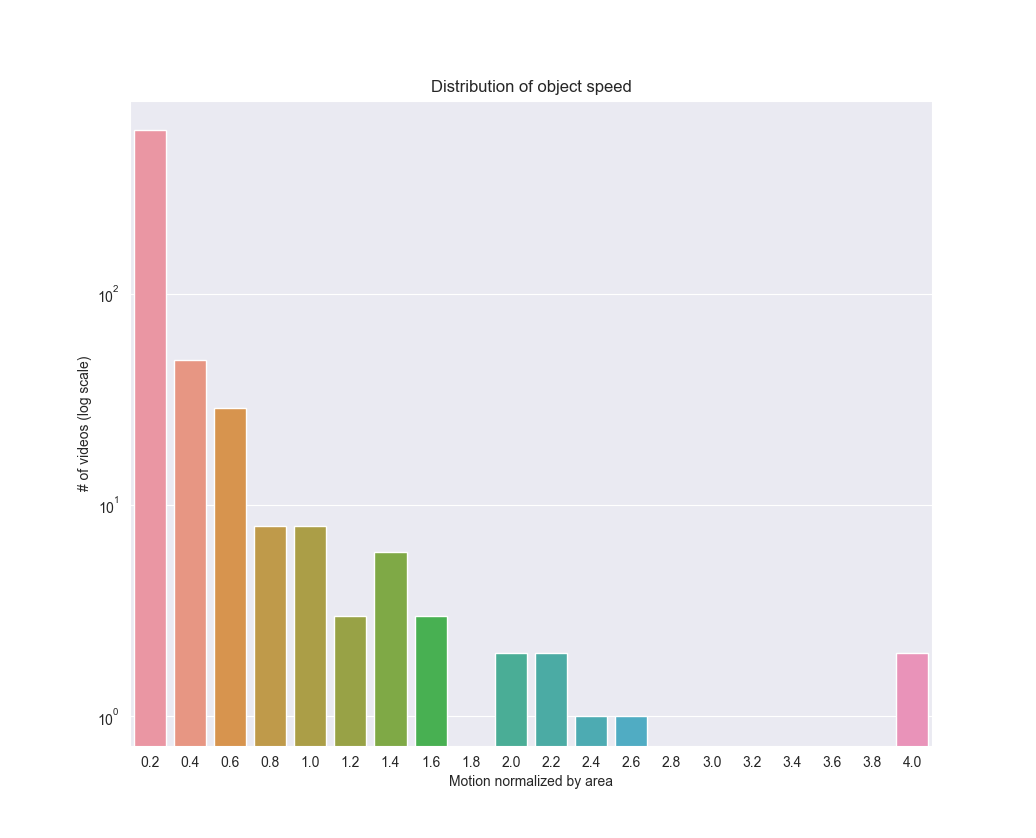}
\end{center}
\vspace{-7mm}
  \caption{
  Distribution of object motion normalized by the object area in VOST. Most videos are smooth but there is a significant amount of challenging sequences with fast motion.
  }
\vspace{-3mm}
\label{fig:spd}
\end{figure}

Finally, in Figure~\ref{fig:spd} we show the distribution of object motion at 5 fps, proportional to the
size of the object. To this end, we follow~\cite{dave2020tao} and compute the distance between the centers of bounding boxes enclosing the objects mask in frames $t$ and $t-1$ in the horizontal dimension as $d^t_x=\frac{||x_{t-1} - x_t||}{a_{t-1}}$, where $a_{t-1}$ is the bounding box area in frame $t-1$. The distance in the vertical dimension $d^t_y$ is computed in the same way, and the combined distance is $d_t=||d_x^t,d_y^t||_2^2$. We plot the largest motion in each video and observe that while most videos are temporally smooth there is a significant number of clips with fast motion, which often correspond to the object going out of frame. As we saw in Table~\ref{tab:abl} in the main paper, such sequences are especially challenging for existing VOS algorithms.

\begin{figure*}[t]
    \centering
    \includegraphics[width=1\linewidth]{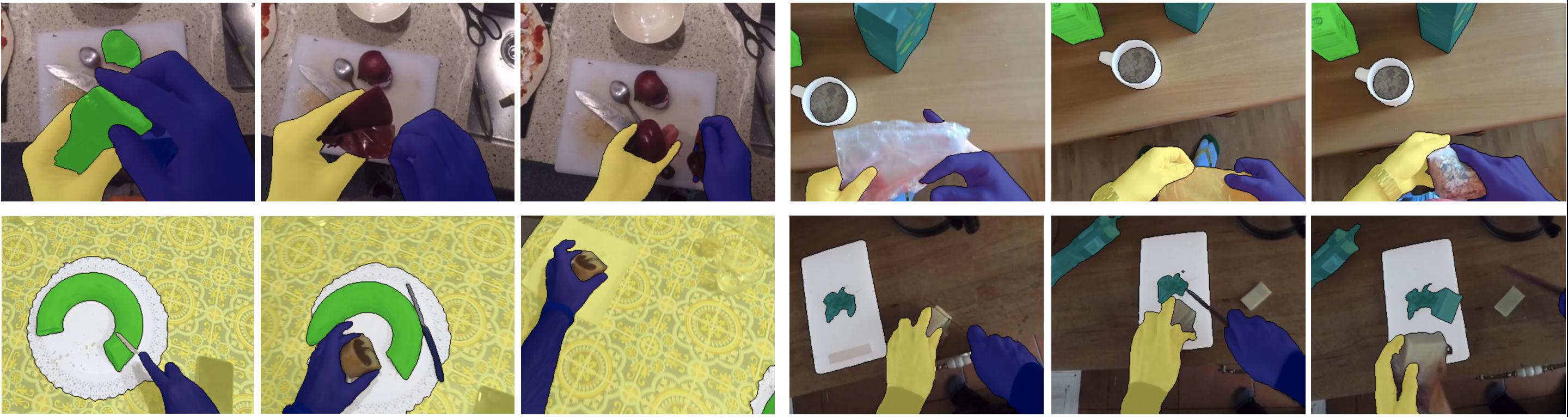}
    \vspace{-5mm}
    \caption{Visualization of automatically interpolated and filtered VISOR labels~\cite{darkhalil2022epic}. Colours indicate instance ids. We can see that automatic interpolation fails during transformations, such as peeling of the onion or folding of the cereal bag in the top row, whereas the objects with stable appearance, such as hands or boxes on the table are perfectly segmented. In VOST we are focusing precisely on the scenarios that automatic interpolation cannot handle, justifying our decision to densely label videos at 5 fps. }
    \label{fig:visor}
    \vspace{-3mm}
\end{figure*}
\section{Limitations of Annotation Interpolation}
\label{sec:interp}
Several works have recently proposed to scale the size of VOS datasets by labeling at a very low fps and then interpolating ground truth labels to obtain temporally dense masks~\cite{darkhalil2022epic,wang2021unidentified}. As interpolation can fail, they automatically filter out the unreliable results and only keep the accurate trajectories. We now demonstrate that this approach fails precisely for the objects that undergo non-trivial transformations, justifying our decision to label VOST at 5 fps.

To this end, we visualize the interpolated labels from VISOR~\cite{darkhalil2022epic} for some of the sequences that feature object transformations in Figure~\ref{fig:visor}. In the first video with onion peeling we can see that the interpolation fails as soon as the object starts to transform, with only the actor's hands accurately segmented. In the bag folding video in the top right interpolation succeeds in the middle of the sequence, but fails at the more challenging earlier and later parts. Note that the static boxes on the table, on the other hand, are perfectly segmented for the entire duration of the video. The cake cutting example in the bottom left of Figure~\ref{fig:visor} illustrates how interpolation fails to capture the part of the object that is separated from the rest of the cake. Finally, in the cheese cutting example in the bottom right interpolation fails for the entire duration of the sequence. Moreover, a part of the cheese is merged with the vegetables on the cutting board at the end. In contrast, VOST provides accurate, temporally dense labels even for the most challenging sequences.

\section{Details of the R-STM architecture}
\label{sec:rstm}
We first provide a brief overview of the Long Short-Term Transformer (LSTT) architecture used in AOT~\cite{yang2021associating}, which we extend with a recurrent transformer module. We omit some of the unimportant details of LSTT architecture for brevity. Please see the original paper for a full description. 

As the name suggests, LSTT combines two attention modules, $AttLT$ and $AttST$, that are implemented as transformers and are used to query long- and short-term memory respectively. Concretely,
\begin{equation}
    AttLT(F^t, M^{t}) = Att(F^t W^Q, M^{t} W^K, M^{t} W^V),
\end{equation}
where $F^t$ is the feature encoding of the current frame, $M^{t}$ is the memory state, $Att$ is the standard, multi-head attention operation~\cite{vaswani2017}, and $W^Q, W^K, W^V$ are linear projections.
Crucially, the memory state $M^{t}$ is simply a concatenation of per-frame feature maps from previous $N$ time-steps: $M=Concat(F^{t_1}, F^{t_2}, ..., F^{t_N})$ combined with corresponding instance segmentation maps (either ground truth or predicted by the model). Short term memory is defined in the same way, with the main difference being that $N$ is fixed to 1 in practice, so, effectively:
\begin{equation}
    AttST = AttLT(F^t, F^{t-1}).
\end{equation}
The outputs of both attention operations are then summed and the result is used to decode the instance masks of the target objects in the current frame.

It is easy to see that these attention operations perform appearance-based patch retrieval as the frame-level feature maps $F$ can only encode, static appearance information. This is in stark contrast to traditional spatio-temporal memory modules~\cite{ballas2015delving,shi2015convolutional} that feature a single memory state tensor that is recurrently updated and can thus aggregate relevant information from the entire video. This is not only more computationally efficient than stacking feature maps, but also allows to represent concepts that are not explicitly present in any of the frames (e.g. locations of occluded objects).

To incorporate this capability into LSTT, we replace the short-term memory  with a recurrent transformer ($R\mbox{-}STM$):
\begin{equation}
    R\mbox{-}STM(F^t, M^{t}) = Att(F^t W^Q, K, V),
\end{equation}
where $K=norm(W^K_F F^{t}+W^K_M M^{t})$, similarly $V = norm(W^V_F F^{t}+ W^V_M M^{t})$, and $norm$ denotes layer normalization~\cite{ba2016layer}. Crucially, $M^{t+1} = R\mbox{-}STM(F^t, M^{t})$ making it a recurrent memory module. Our experiments in Table~\ref{tab:arch} in the main paper demonstrate that this simple modification indeed improves the transformation modeling capacity of AOT, but a more comprehensive approach for modeling spatio-temporal information is required to fully address the problem.

\section{Further Implementation Details}
\label{sec:impl}
All the models are trained and evaluated at 5 fps unless stated otherwise. When fine-tuning on VISOR~\cite{darkhalil2022epic} we excluded EPIC-KITCHENS~\cite{damen2022rescaling} videos that were used in the validate or test sets of VOST. For AOT~\cite{yang2021associating} and AOT+ we use the R50-L variant of the model and replace their default strategy of adding every fifth frame to the long term memory at inference time, which does not scale to long videos, with sparse insertion strategy proposed in~\cite{cheng2022xmem}. We found that training CRW~\cite{jabri2020space} at a higher $512 \times 512$ resolution leads to a slightly improved performance on VOST so we follow this strategy in our experiments. We also found that all baselines treat “Ignore” as another instance label. We modified their implementations to skip ignored regions in the first frame of a sequence and not include these pixels in the loss computation. Otherwise we left the original implementations and hyper-parameters unchanged for all of the methods, only adjusting the number of fine-tuning iterations on the validation set of VOST.

\newpage

\end{document}